\documentclass[sigconf,balance=false]{acmart}
\usepackage{popets}
\usepackage[inline]{enumitem}
\usepackage{dirtytalk}
\usepackage{booktabs}
\usepackage{amsmath,amsfonts}
\usepackage{algorithmic}
\usepackage{graphicx}
\usepackage{textcomp}
\usepackage{placeins}   
\usepackage{pgffor} 
\usepackage{multirow}
\usepackage[table,xcdraw]{xcolor}
\usepackage{booktabs}
\usepackage{subcaption}
\usepackage{natbib}
\newtheorem{definition}{Definition}
\newtheorem{proposition}{Proposition}

\def \baseline{\texttt{Baseline}}
\def \dponly{\texttt{DP-only}}
\def \faironly{\texttt{Fair-only}}
\def \dpfair{\texttt{DP+Fair}}

\def \eg{\emph{e.g.}}
\def \ie{\emph{i.e.}}

\def \data{$D$}

\setcopyright{popets}
\copyrightyear{2026}
\acmYear{2026}
\acmVolume{2026}
\acmNumber{4}
\acmDOI{10.56553/popets-2026-0110}
\acmISBN{}
\acmConference{Proceedings on Privacy Enhancing Technologies}
\begin{document}
\sloppy
\title{Where to Intervene? Benchmarking Fairness-Aware Learning on Differentially Private Synthetic Tabular Data}

%%%%%%%%%%%%%%%% Authors' Info %%%%%%%%%%%%%%%%%

\author{Vinícius Gabriel Angelozzi}
\orcid{0009-0000-4049-0401}
\affiliation{%
  \institution{Centre Inria de l'Université Grenoble Alpes}
  \city{}
  \country{}}
\email{verona.projects@tutanota.com}

\author{Héber H. Arcolezi}
\orcid{0000-0001-8059-7094}
\affiliation{%
  \institution{ÉTS Montréal, Inria Grenoble}
  \city{}
  \country{}}
\email{heber.hwang-arcolezi@etsmtl.ca}


\renewcommand{\shortauthors}{V.G. Angelozzi \& H.H. Arcolezi}

\begin{abstract}
Machine learning models are increasingly deployed in high-stakes domains, raising concerns about both privacy and fairness. Differential Privacy (DP) has become a gold standard for privacy-preserving data analysis, while fairness-aware mechanisms aim to mitigate discrimination against underrepresented groups. However, these objectives can conflict: DP often amplifies disparities across demographic groups, and little is known about whether established fairness interventions remain effective under DP constraints. In this work, we present, to our knowledge, the first systematic evaluation of fairness interventions on differentially private synthetic tabular data. Our benchmark centers on the \emph{Adaptive Iterative Mechanism (AIM)}, identified as the state-of-the-art marginal-based DP synthesizer in recent KDD \& VLDB 2025 tutorials by Cormode et al.~\cite{Cormode2025}. We thus evaluate fairness interventions across four datasets, multiple group fairness metrics, and three categories of mitigation strategies (pre-processing, in-processing, and post-processing) under a wide range of privacy budgets. We compare four pipeline configurations: (\emph{Baseline}) training on original data; (\emph{DP-only}) training on DP synthetic data; (\emph{Fair-only}) applying fairness mechanisms on original data; and (\emph{DP+Fair}) combining fairness mechanisms with DP synthetic data. Our results demonstrate that while DP alone can degrade both utility and fairness, applying fairness interventions can partially restore equitable outcomes. Among them, \emph{post-processing methods tend to provide more stable fairness–utility trade-offs across privacy budgets and synthesizers}, achieving strong fairness improvements while preserving competitive utility relative to other intervention stages. We release all code, data, and experimental artifacts in an open-source repository (\url{https://github.com/vinicius-verona/dp-fair-intervention-benchmark}) to ensure full reproducibility and to support future research on the privacy-fairness-utility trade-off.
\end{abstract}


\maketitle

\section{Introduction}

Synthetic tabular data generation is increasingly adopted by the research community~\cite{Hu2024,jordon2022synthetic,liu2024tabular}, regulators~\cite{fsa,cnil}, and industry~\cite{tumult_labs,ydata} as a promising approach to facilitate data sharing and downstream analysis while reducing disclosure risks. 
These methods aim to approximate the empirical distribution of real datasets and generate artificial records that preserve key statistical properties. 
However, synthetic data does not guarantee privacy or fairness by default. 
On the one hand, without formal protections, generative models may memorize or leak sensitive information from the training data, making them vulnerable to membership inference, attribute inference, or reconstruction attacks~\cite{Ganev2025,yao2025dcr,Stadler2022,giomi2023unified}.
Moreover, synthetic data can replicate or even amplify societal biases present in the data, leading to unfair outcomes in downstream predictive models~\cite{Barocas2023}.

Differential privacy (DP)~\cite{Dwork2006Calibrating,dp:Cynthia2014} provides rigorous, quantifiable protection by bounding the influence of any single record on model training or statistics. 
DP-based synthetic data generators~\cite{dp:aim,Zhang2017,torkzadehmahani2019dp} address the privacy gap by injecting calibrated noise during training or sampling. 
Most recent benchmarks have therefore concentrated on the \emph{utility} of DP synthetic data, evaluating predictive performance across generative models, tasks, and privacy levels~\cite{tao2021benchmarking,Rosenblatt2023,qian2023synthcity,Ganev2024}. 
Yet, beyond aggregate utility, the statistical distortions introduced by formal privacy guarantees may interact with subgroup structure in non-trivial ways.

Beyond utility, however, the interplay between DP and fairness has received far less attention, and is often more complex~\cite{yao2025sok,fair-dp:fioretto2022}. 
When DP is applied directly to model training (\eg, via DP-SGD~\cite{Abadi2016} or PATE~\cite{papernot2018scalable}), prior work has shown that noise reduces overall accuracy but disproportionately harms minority or underrepresented groups~\cite{dp:Eugene2019}. 
This phenomenon, often described as \emph{DP disparate impact}, raises the question of whether similar effects also emerge in models trained on DP synthetic data.

Recent studies confirm that they do. 
Work on \emph{DP synthetic data}~\cite{bullwinkel2022evaluating,dp:ganev2022,Pereira2024} has shown that privacy mechanisms and data characteristics interact in subtle ways, often leading to heterogeneous fairness outcomes across groups. 
However, these evaluations are largely observational: they measure disparities (\eg, statistical parity or subgroup accuracy) but do not examine whether \emph{fairness-aware learning mechanisms} remain effective when applied to DP synthetic data.

This gap motivates our central research question: \textbf{\textit{``Where should one intervene in the machine learning (ML) pipeline to mitigate unfairness under DP synthetic data?''}} 
Should interventions target the data distribution (pre-processing), the training procedure (in-processing), or the model outputs (post-processing)? 
More fundamentally, how do formal differential privacy guarantees reshape the stability, effectiveness, and relative competitiveness of these intervention strategies across privacy budgets?

The question is timely. 
Several commercial platforms now offer \emph{DP synthetic data} as a product~\cite[Table 1]{Ganev2025}, including vendors such as Tumult Labs~\cite{tumult_labs} and YData~\cite{ydata}. 
These platforms are already being used in sensitive domains such as healthcare and finance. 
As fairness concerns grow, both ethically and legally (\eg, EU AI Act~\cite{AIACT2024}), developers may need to integrate fairness-aware interventions \emph{without redesigning entire DP generation pipelines}.

\textbf{Contributions.}
Building on these observations, we conduct a structured benchmarking study of fairness mitigation strategies in DP synthetic data pipelines.
Our goal is to characterize when and how fairness-aware mechanisms can counteract disparate impact introduced by differential privacy, while preserving predictive utility under explicit privacy budgets.
To ensure relevance and rigor, we center our study on the \emph{Adaptive Iterative Mechanism (AIM)}~\cite{dp:aim}, recognized as the current \emph{state-of-the-art differentially private synthesizer for tabular data} in recent \emph{KDD \& VLDB 2025 tutorials by Cormode et al.~\cite{Cormode2025}} and \emph{utility-driven} benchmarks~\cite{tao2021benchmarking,Rosenblatt2023,qian2023synthcity,Ganev2024}.
For completeness, we also report results with the \emph{Maximum Spanning Tree (MST)}~\cite{dp:mst} synthesizer, winner of the NIST Differential Privacy Synthetic Data Challenge, in Appendix~\ref{app:add_results}. 
By focusing on widely adopted marginal-based DP generators, our study isolates the interaction between formal privacy guarantees and downstream fairness mitigation.
In summary, our main contributions are:

\begin{itemize}[leftmargin=*]
    \item We present, to our knowledge, the first systematic evaluation of fairness interventions applied to classifiers trained on DP synthetic tabular data, explicitly analyzing how differential privacy reshapes fairness--utility trade-offs.
    An overview of our benchmark pipeline is shown in Figure~\ref{fig:methodology}.
    
    \item We evaluate pre-, in-, and post-processing fairness interventions under varying privacy budgets, revealing how the effectiveness and stability of these mechanisms change when applied to DP synthetic rather than original data.
    
    \item Using multiple real-world datasets, three ML classifiers, and two state-of-the-art DP synthesizers, we characterize how DP alters fairness--utility dynamics and identify conditions under which interventions are more or less effective at mitigating unfairness.
    
    \item We release all code, datasets, and experimental artifacts to support reproducibility and enable future research at the intersection of privacy and fairness (see Section~\ref{sub:implementation} and the \textbf{GitHub}: \url{https://github.com/vinicius-verona/dp-fair-intervention-benchmark}).
\end{itemize}

\textbf{Findings.}  
Our results show that fairness interventions applied to models trained on DP synthetic data can partially recover fairness degradation induced by DP noise, although the effect is metric-, budget-, and stage-dependent. 
Across datasets, privacy budgets, utility metrics, and learning models, Reweighing (RW)~\cite{fair:Reweighing-paper}, Reject Option Classification (ROC)~\cite{kamiran2012decision}, and Equalized Odds Post-Processing (EqOdds)~\cite{fair:hardt2016equality} emerge as the most reliable interventions. 
RW provides stable fairness improvements, especially for parity-oriented metrics, but often incurs measurable utility loss. 
In contrast, ROC and EqOdds tend to offer the strongest fairness--utility trade-offs, frequently appearing on or near the Pareto frontier, particularly for EOD- and SPD-oriented analyses. 
In-processing mechanisms, especially EGR, preserve utility close to DP-only but achieve more limited disparity reductions in this DP synthetic setting.

\section{Related Work} \label{sec:related_work}

\textbf{Privacy and fairness.} 
The interaction between privacy and fairness has been widely studied~\cite{makhlouf2024systematic,makhlouf2024impact,dp:Eugene2019,fair:Chang2021,Ganev2024,arcolezi2025fair,aalmoes2022leveraging,uniyal2021dp,farrand2020neither}. 
A consistent finding is that differentially private learning (\ie, through DP-SGD~\cite{Abadi2016} or PATE~\cite{papernot2018scalable}) can exacerbate disparities across demographic groups, a phenomenon often referred to as the \emph{disparate impact of DP}. 
For instance,~\cite{dp:Eugene2019} shows that DP-SGD reduces overall accuracy but disproportionately harms minority groups. 
Surveys such as~\cite{fair-dp:fioretto2022,yao2025sok} further highlight that privacy and fairness objectives may align in some regimes but conflict in others, depending on the data distribution and model class.
These studies primarily analyze fairness when DP is applied directly to model training.

\textbf{DP synthetic data and utility.} 
The release of \emph{DP synthetic data} has motivated a growing line of work benchmarking the \emph{utility} of different generative models. 
Early studies compare marginal-based synthesizers (\eg, AIM~\cite{dp:aim}, MST~\cite{dp:mst}) with deep generative approaches (\eg, DP-GANs~\cite{torkzadehmahani2019dp,xu2019modeling}), showing that marginal-based methods often yield higher utility on tabular datasets~\cite{tao2021benchmarking,Rosenblatt2023,qian2023synthcity,Ganev2024}. 
These results position marginal-based models as competitive baselines for structured domains.

\textbf{DP synthetic data and fairness.} 
More recent work evaluates fairness explicitly. 
\cite{bullwinkel2022evaluating} compares four DP synthesizers across multiple datasets and privacy budgets, finding that most degrade fairness but that MST behaves more favorably than GAN-based alternatives. 
Similarly,~\cite{Pereira2024} analyzes fairness and utility metrics for both GAN-based and marginal-based synthesizers, showing that the latter tend to preserve subgroup accuracy and often maintain or improve group fairness metrics. 
The work in~\cite{dp:ganev2022} provides a fine-grained analysis of subgroup disparities in DP synthetic data, showing that DP can either amplify or mitigate imbalance depending on the model, and that classifiers trained on such data exhibit reduced performance for minority groups.
These studies demonstrate that privacy-preserving data generation can reshape fairness properties, but they remain largely observational.

\textbf{Positioning of our work.} 
Prior works on fairness in DP synthetic data~\cite{bullwinkel2022evaluating,Pereira2024,dp:ganev2022} primarily measure how differential privacy affects fairness metrics after data release.
They do not systematically analyze how \emph{fairness-aware intervention mechanisms} behave when deployed on top of DP synthetic data under varying privacy budgets.
In particular, the interaction between intervention stages (pre-, in-, or post-processing) and formal privacy constraints has not been jointly characterized in a unified empirical setting.
\emph{Our work addresses this gap} by studying fairness mitigation strategies applied \emph{after} DP synthetic data generation, explicitly isolating the downstream effects of formal privacy guarantees on fairness–utility trade-offs.
Rather than designing jointly fair-and-private synthesizers (\eg,~\cite{PFGuard}), we focus on the practical and increasingly common deployment scenario in which DP synthetic data is produced first, and fairness considerations are addressed subsequently.
This separation allows us to characterize how established fairness mechanisms behave under fixed privacy budgets, and to identify which intervention stages remain stable or degrade under privacy constraints.

\section{Preliminaries} \label{sec:background}

In this section, we briefly review about generative models, differential privacy, and fairness-aware learning.

\subsection{Generative Models and Synthetic Data} \label{sub:generative_models}

Generative models aim to approximate the distribution of real data and to produce artificial records that preserve its statistical properties. 
Let \(D = \{x_i\}_{i=1}^n\), with \(x_i \in \mathcal{X}\), denote the original dataset drawn \emph{i.i.d.} from an unknown data-generating distribution \(p_{\text{data}}\). 
A generative model learns a parameterized distribution \(p_\theta\) intended to approximate \(p_{\text{data}}\). 
Synthetic records are then generated by sampling \(\tilde{x}_j \sim p_\theta\), yielding a synthetic dataset \(\tilde{D} = \{\tilde{x}_j\}_{j=1}^m\) that is statistically similar to \(D\). 
This synthetic dataset is released in place of the original data for downstream analysis, with the goal of enabling utility while mitigating disclosure risk.

While many approaches exist for building generative models, in this work we focus on marginal-based methods~\cite{dp:aim,dp:mst}. 
These synthesizers are rooted in Bayesian network formulations that decompose the joint distribution of tabular data into lower-dimensional marginals and conditional dependencies. 
Such approaches have consistently shown strong performance on structured tabular data~\cite{tao2021benchmarking,Ganev2024}, where feature dependencies can be effectively captured by explicit probabilistic modeling. 
In contrast, deep generative models, while highly successful for image or text synthesis, often face challenges in faithfully modeling heterogeneous tabular data. 

\subsection{Differential Privacy} \label{sub:dp}

Differential privacy is a property of randomized mechanisms that limits how much the output distribution can change when a single individual’s record is modified. 
Intuitively, it enables learning about the population while revealing little about any one person~\cite{dp:Cynthia2014}. 
We adopt the standard \(\varepsilon\)-DP notion~\cite{Dwork2006Calibrating}.

\begin{definition}[\(\varepsilon\)-Differential Privacy]
Let \(\mathcal{X}\) be the data domain and let datasets \(D,D' \in \mathcal{X}^n\) be \emph{neighbors} (written \(D \sim D'\)) if they differ in exactly one individual’s record. Let \(\mathcal{O}\) denote the output space of a randomized mechanism. A randomized mechanism \(\mathcal{M}:\mathcal{X}^n \to \mathcal{O}\) satisfies \(\varepsilon\)-DP if for all measurable \(S \subseteq \mathcal{O}\),
\[
\Pr[\mathcal{M}(D)\in S] \le e^{\varepsilon}\, \Pr[\mathcal{M}(D')\in S].
\]
\end{definition}

When \(\varepsilon=0\), outputs for neighboring datasets are identically distributed, implying that the mechanism’s output cannot depend on any single record. 
Larger \(\varepsilon\) permits greater sensitivity to an individual, weakening privacy. 
Thus, choosing \(\varepsilon\) entails a privacy-utility trade-off. 
We now recall the post-processing property of differential privacy\footnote{For clarity, we use "DP post-processing" when referring to the privacy property, and "fairness post-processing" when referring to mitigation strategies.}, which is central to our deployment model.

\begin{proposition}[DP Post-Processing Property~\cite{dp:Cynthia2014}]
Let $\mathcal{M}:\mathcal{X}^n \to \mathcal{O}$ be an $\varepsilon$-differentially private mechanism, and let $f:\mathcal{O}\to\mathcal{O}'$ be any (possibly randomized) mapping that does not access the original dataset $D$.
Then the composed mechanism $f \circ \mathcal{M}$ is also $\varepsilon$-differentially private.
\end{proposition}
This property ensures that once a DP synthetic dataset $\tilde{D}=\mathcal{M}(D)$ is released, any downstream learning procedure or fairness intervention that operates solely on $\tilde{D}$ constitutes pure DP post-processing and cannot weaken the privacy guarantee with respect to the original training records $D$.

\subsection{Fairness-Aware Learning} \label{sub:fair_learning}

Fairness-aware learning aims to incorporate fairness criteria into predictive models. 
Let \(X\) denote features, \(A\in\{0,1\}\) a protected attribute, \(Y\in\{0,1\}\) the label, and \(h:\mathcal{X}\!\to\!\{0,1\}\) (or score \(s:\mathcal{X}\!\to\![0,1]\)) the predictor. 
Models may exhibit biased behavior for diverse reasons: some biases are intrinsic to the data (data-to-model bias), while others emerge from missing representative samples or from limitations and objectives of the learning algorithm~\cite{Barocas2023, fair:review}. 
To address these issues, algorithmic interventions are typically organized by \emph{where} they act in the pipeline:

\begin{itemize}[leftmargin=*]
    \item \textbf{Pre-processing.} Methods that transform the training data \(D\) into \(\hat{D}\) (\eg, reweighting or data transformation) to reduce dependence on the protected attribute \(A\) while preserving task utility under the defined fairness notion.
    
    \item \textbf{In-processing.} Methods that modify the learning algorithm or objective. 
    Let \(\mathcal{L}(h;D)\) denote a task-specific empirical loss function evaluated on dataset \(D\), and let \(\tau \ge 0\) be a user-defined fairness tolerance. 
    In-processing approaches either solve a constrained optimization problem
    \(
    \min_{h}\ \mathcal{L}(h;D)\quad \text{s.t.}\quad \Delta_{\mathrm{fair}}(h)\le \tau,
    \)
    or incorporate fairness directly into the objective via a penalty term,
    \(
    \mathcal{L}(h;D)+\lambda\,\Omega_{\mathrm{fair}}(h),
    \)
    where \(\Omega_{\mathrm{fair}}(h)\) measures unfairness and \(\lambda \ge 0\) controls the fairness-utility trade-off.

    \item \textbf{Post-processing.} Methods that adjust predictions of a trained model (\eg, group-specific thresholds, calibration, or randomized decisions) to satisfy target constraints with minimal utility loss, treating the model as a black box.
\end{itemize}

\section{Benchmark Design} \label{sec:methodology}
In this section, we first describe the overall benchmark structure and experimental configurations, followed by details on datasets and prediction tasks, the DP generative models used, the fairness mechanisms evaluated, as well as the model training procedure, including data preprocessing, classifier configuration, and protocol for stability and reproducibility.

\subsection{Overview and Experimental Configuration} \label{sub:overview_benchmark}

\begin{figure*}
    \centering
    \includegraphics[width=1\linewidth]{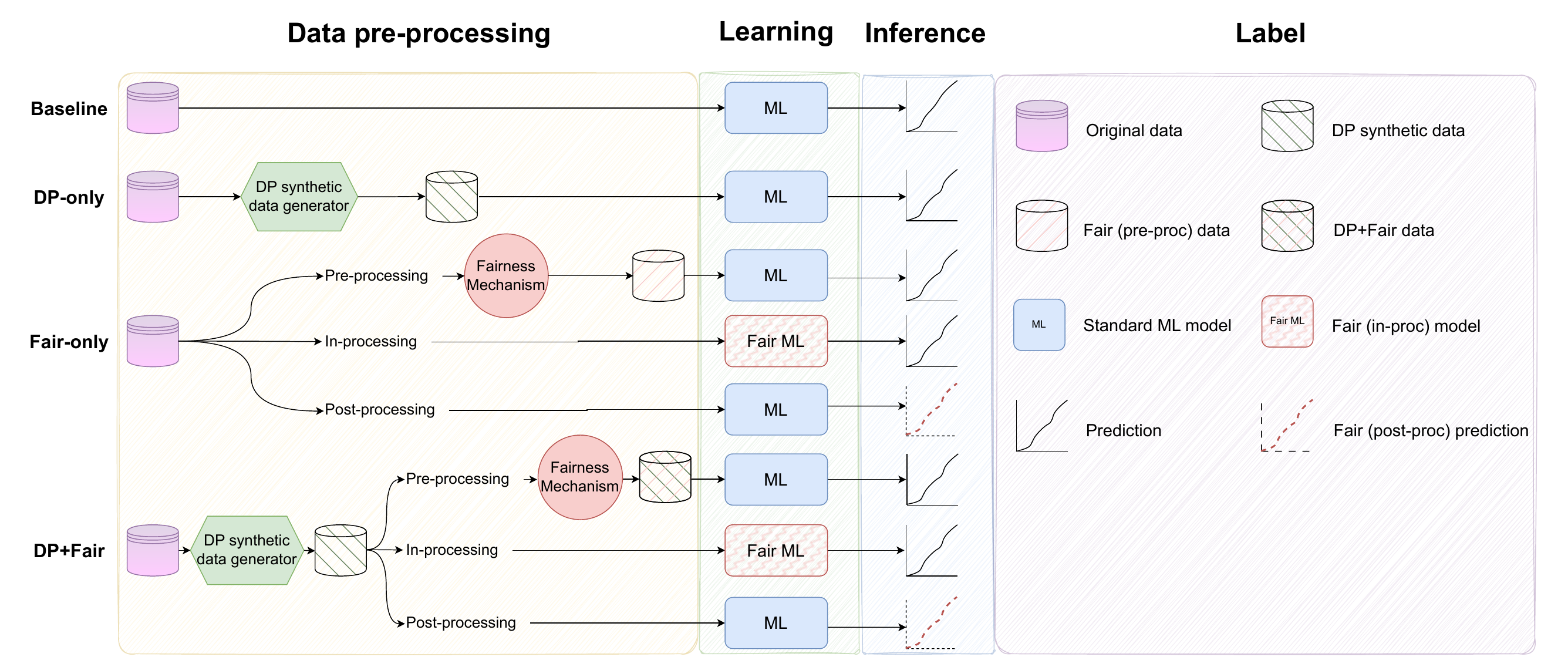}
    \caption{Overview of our benchmark design. 
    We evaluate fairness-aware learning mechanisms applied at three intervention stages, pre-processing, in-processing, and post-processing, across four configurations: 
    (1) \baseline{} (original data, no fairness intervention), 
    (2) \dponly{} (DP synthetic data, no fairness intervention), 
    (3) \faironly{} (original data with fairness intervention), and 
    (4) \dpfair{} (fairness interventions on DP synthetic data). 
    This design systematically captures the isolated and combined effects of privacy and fairness interventions.} 
    \label{fig:methodology}
\end{figure*}

Figure~\ref{fig:methodology} illustrates the overall benchmark pipeline. 
Starting from the original dataset, we construct four prediction pipelines, corresponding to the configurations described in the following. 
Each pipeline follows the same three stages, namely, data pre-processing, model training, and inference, but differs in whether DP is applied to the data and whether fairness interventions are applied before, during, or after training. 
This setup enables a systematic comparison of privacy, fairness, and utility across intervention points.

\begin{enumerate}[leftmargin=*]

    \item \textbf{\baseline.} A standard machine learning model trained directly on the original (non-private, non-fair) training data \data. 
    This serves as the \emph{reference point} to evaluate the impact of privacy and fairness interventions.
    
    \item \textbf{\dponly.} The original training data \(D\) is replaced with \emph{differentially private synthetic data} \(\tilde{D}\). 
    No fairness mechanism is applied. 
    This setting isolates the effect of differential privacy on model performance and fairness metrics.
    
    \item \textbf{\faironly.} A fairness intervention is applied to the original training data, to the learning algorithm, or to the model output, without incorporating any differential privacy. 
    Specifically, we consider three families of interventions (see Section~\ref{sub:fair_learning}): \emph{pre-processing}, \emph{in-processing}, and \emph{post-processing}. 
    This setting quantifies the effect of fairness mechanisms in isolation.    
    
    \item \textbf{\dpfair.} Privacy and fairness interventions are combined. 
    A fairness mechanism is applied either (i) to the DP synthetic data before training (pre-processing), (ii) during model training (in-processing), or (iii) to the model predictions (post-processing). 
    This setting evaluates whether fairness mechanisms remain effective when operating on DP synthetic data, and whether they can mitigate the fairness degradation introduced by DP.
\end{enumerate}

\subsection{Data and Task} \label{sub:datasets}

\paragraph{Dataset} We conduct our benchmark on four open datasets widely used in fairness research: 

\begin{itemize}[leftmargin=*]
    \item \textbf{Adult}~\cite{dataset:adult_2} (UCI Census Income) contains $n = 47,621$ individuals and the goal is to predict whether a person's income exceeds \$50K/year based on 10 demographic and occupational attributes. 
    \emph{Gender} is used as the protected attribute for fairness evaluation.    
    
    \item \textbf{COMPAS}~\cite{compas} includes $n = 5,050$ defendants and the goal is to predict recidivism risk based on 7 criminal history and demographic attributes. 
    \emph{Race} is used for fairness evaluation.
    
    \item \textbf{ACSIncome}~\cite{ding2021retiring} extends Adult with richer socioeconomic features from the U.S. Census American Community Survey. 
    We select the Utah state subset with $n = 16,337$ individuals.
    We set the income threshold at the median value ($>38$K/year) and use \emph{gender} as the protected attribute for fairness evaluation.
    
    \item \textbf{BiasOnDemand}~\cite{bias_on_demand} is a synthetic dataset generator designed to benchmark fairness and bias under controlled conditions. 
    We use it to simulate data distributions with known levels of group imbalance and label bias. In total, 6 bias configurations were tested across 3 categories (see Table~\ref{tab:BoD} in Appendix~\ref{app:datasets}: imbalance, historical bias, and measurement bias. 
    By default, we present the results with \texttt{Config 5} in the full paper.
    In total, we generate $n = 30,000$ samples, and the goal is to predict the binary value $Y$ conditioned on 2 attributes. 
    The configurations studied are set in a way to:
    \begin{enumerate*}[label=(\roman*)]
        \item isolate bias on target;
        \item add imbalance to the dataset;
        \item isolate bias to a feature;
        \item add bias to a feature and increase feature correlation and dependence.
    \end{enumerate*}
    Furthermore, the values of each configuration parameter were set by experimenting with different values until high values of fairness metrics were achieved, indicating strong bias.
\end{itemize}

\paragraph{Task} 
All datasets are cast as binary classification tasks.
This choice reflects the dominant and most fundamental and mature experimental setting in the fair machine learning literature, where the majority of group fairness definitions (\eg, Statistical Parity~\cite{dwork2011fairnessawareness}, Equal Opportunity~\cite{fair:hardt2016equality}) and corresponding mitigation mechanisms (\eg, Calibrated Equalized Odds Post-processing~\cite{fair:CEOP}, and Reweighing~\cite{fair:Reweighing-paper}) are formally defined
and/or empirically validated for binary labels and binary protected attributes.
Moreover, our benchmark is implemented on top of AIF360~\cite{fair:AIF360}, which natively supports binary classification and binary protected attributes for all provided fairness interventions, enabling consistent and reproducible comparisons across methods. 
In addition, some of these provided interventions do not offer the same support to multi-label classification. Exponentiated Gradient Reduction~\cite{fair:Reduction} and Grid Search Reduction~\cite{fair:Reduction}, for example, require an estimator as a parameter, and have the following constraint: \say{labels \textit{y} and predictions returned by \textit{predict(X)} are either 0 or 1}. 
Such restrictions, together with the necessity to standardize our experiments and the fact that multi-label classifications can be reduced to a binary classification problem, lead us to proceed with a first experimental setup based on the binary classification tasks.
Accordingly, we standardize continuous features, binarize the protected attribute \(A \in \{0,1\}\), and encode the target variable as binary \(Y \in \{0,1\}\).
Further details on dataset preprocessing are provided in Appendix~\ref{app:datasets}.

\subsection{DP Generative Models}  \label{sub:dp_models}

To generate differentially private synthetic data, we focus on two \emph{marginal-based synthesizers}, AIM~\cite{dp:aim} and MST~\cite{dp:mst}, described hereafter, both implemented in the SmartNoise library (\href{https://docs.smartnoise.org/}{https://docs.smartnoise.org/}).
Our choice is motivated by recent utility-oriented benchmarks~\cite{tao2021benchmarking,Rosenblatt2023,qian2023synthcity,Ganev2024}, which consistently show that marginal-based models outperform deep generative approaches such as DP-GANs~\cite{torkzadehmahani2019dp} on tabular data. 
Whereas deep generative models excel in unstructured domains like images, they often struggle with the heterogeneous feature types and sparsity typical of structured tabular data. 
By contrast, marginal-based synthesizers explicitly model low-dimensional marginals and conditional dependencies, leading to higher fidelity and stronger predictive performance in downstream classification tasks. 
We therefore adopt AIM and MST as representative DP synthetic data generators.

\begin{itemize}[leftmargin=*]
    \item \textbf{Adaptive Iterative Mechanism (\textbf{AIM})~\cite{dp:aim}.} AIM is a state-of-the-art differentially private synthetic data generation algorithm. It follows a select-measure-generate paradigm: it iteratively selects informative sets of marginal queries, measures them under differential privacy by adding calibrated noise, and uses Private-PGM~\cite{mckenna2019graphical} to infer an approximate joint distribution from the noisy measurements. 
    Synthetic records are then sampled from this inferred distribution.

    \item \textbf{Maximum Spanning Tree (\textbf{MST})~\cite{dp:mst}.} MST also follows a select-measure-generate paradigm. 
    It constructs a dependency graph over features using privately measured pairwise correlations, selects a maximum spanning tree to determine which low-dimensional marginals to measure, and then measures these marginals under DP. 
    As in AIM, the resulting noisy marginals are passed to Private-PGM~\cite{mckenna2019graphical}, which estimates an approximate joint distribution and samples synthetic records from it.
\end{itemize}

Our focus on AIM and MST is, therefore, intentional rather than exhaustive. 
These methods represent the family of marginal-based DP synthesizers that have repeatedly been shown to be competitive for structured tabular data, and they provide a controlled setting for studying how DP synthetic data affects downstream fairness interventions. 
Accordingly, our conclusions should be interpreted as applying primarily to high-utility marginal-based DP synthetic data pipelines, rather than to all possible DP generative models.

\subsection{Fairness Mechanisms} \label{sub:fairness_models}
To evaluate whether fairness-aware learning mechanisms remain effective on DP synthetic data, we consider representative methods from the three main categories of interventions: \emph{pre-processing}, \emph{in-processing}, and \emph{post-processing}. 
These methods have been widely studied in the fairness literature and are implemented in open-source libraries such as AIF360~\cite{fair:AIF360}, making them suitable benchmarks for our study.

We deliberately rely on AIF360-based methods because they provide standardized, widely used implementations of canonical fairness interventions across the three main stages of the pipeline. 
This choice allows us to compare intervention stages under a common experimental protocol, while avoiding additional confounding factors due to heterogeneous implementations, incompatible assumptions, or method-specific engineering choices. 
Our goal is therefore not to exhaust the full space of fairness algorithms, but to evaluate representative and reproducible mechanisms that make the stage-level comparison meaningful.

\paragraph{Pre-Processing} These approaches modify the training data before model learning to reduce dependence on the protected attribute.

\begin{itemize}[leftmargin=*]
    \item \textbf{Reweighing (RW)~\cite{fair:Reweighing-paper}.} RW relies on resampling and computing weights for the input samples to decrease discrimination.

    \item \textbf{Disparate Impact Remover (DIR)~\cite{fair:DIR}.} This method transforms the original dataset to reduce disparate impact between privileged and unprivileged subgroups. It first detects whether disparate impact exists, then removes the dependence of unprotected features on the protected feature, and finally adjusts the distributions of unprotected features so that both privileged and unprivileged groups have similar distributions.
    
    \item \textbf{Learning Fair Representations (LFR)~\cite{fair:LFR-paper}.} LFR creates a probabilistic mapping from a data representation in a given input space to a new representation that reduces the ability to identify protected subgroups while preserving task-relevant information. The objective is to approximate statistical parity across groups while balancing fairness with the predictive utility of the data.
\end{itemize}

\paragraph{In-Processing} These methods modify the training procedure itself, typically by solving constrained optimization problems that balance accuracy with fairness.

    \begin{itemize}[leftmargin=*]
        \item \textbf{Exponentiated Gradient Reduction (EGR)~\cite{fair:Reduction}.} This method computes an iterative approximation of the saddle point of a Lagrangian by minimizing the classification loss and maximizing the penalty for fairness violation. The idea behind this computation is the same as a zero-sum game with two players.
         
        \item \textbf{Grid Search Reduction (GSR)~\cite{fair:Reduction}.} This method shares similar ideas with EGR, but relies on brute force. Essentially, it builds a grid of Lagrangian multipliers and exhaustively searches for the best solution considering the fairness-accuracy trade-off.
    \end{itemize}

\paragraph{Post-Processing} These methods operate on the outputs of a trained model, adjusting predictions to satisfy fairness constraints.
\begin{itemize}[leftmargin=*]
    \item \textbf{Reject Option Classification (ROC)~\cite{kamiran2012decision}.} This method operates on the model’s prediction outputs, adjusting decision boundaries to favor fair outcomes in regions of uncertainty. 
    It aims to reduce discrimination by flipping labels for samples near the decision boundary -- particularly when such adjustments enhance fairness for protected groups.
    
    \item \textbf{Equalized Odds Post-Processing (EqOdds)~\cite{fair:hardt2016equality,fair:AIF360}.} This method formulates a linear program to learn probabilities with which the output labels will be changed to satisfy equalized odds constraints while maintaining classification fidelity.
    
    \item \textbf{Calibrated Equalized Odds Post-processing (CEOP)~\cite{fair:CEOP}.} This method operates similarly to EqOdds, enforcing parity in error rates -- false positive rate and false negative rate remain similar across the protected groups. Additionally, CEOP introduces the concern for performing such tasks while maintaining calibration, \ie, ensuring that the predicted scores remain interpretable as true outcome probabilities.
\end{itemize}

\subsection{Metrics} \label{sub:metrics}

\paragraph{Privacy levels} We vary the privacy budget across a broad range, \(\varepsilon \in \{0.05, 0.1, 0.25, 0.5, 0.75, 1, 2, 3, 5, 10, 15, 20\}\), 
covering high-, moderate-, and low-privacy regimes. 
Small values of \(\varepsilon\) provide stronger theoretical privacy guarantees; for example, at \(\varepsilon=0.05\), the likelihood ratio between neighboring datasets is bounded by \(e^{0.05}\approx 1.05\).
Larger values such as \(\varepsilon=10\) or \(20\) correspond to weaker theoretical guarantees, but are included for comparability with prior DP synthetic data benchmarks~\cite{dp:ganev2022,Ganev2024} and because DP mechanisms can still empirically reduce practical attack success compared to non-DP synthetic data~\cite{lowy2024does}.

\paragraph{Utility metrics} We report two standard metrics, namely, accuracy and F1-score (harmonic mean of precision and recall), computed on the held-out test set.

\paragraph{Fairness metrics} Let \(A \in \{0,1\}\) denote the binary protected attribute, \(Y \in \{0,1\}\) the true label, and \(\hat{Y} \in \{0,1\}\) the predicted label. 
We evaluate group fairness using three standard metrics:

\begin{definition}[Model Accuracy Difference (MAD)]
Difference in overall classification accuracy between groups:
\[
\text{MAD} = \Pr[\hat{Y} = Y \mid A = 0] - \Pr[\hat{Y} = Y \mid A = 1] \text{.}
\]
\end{definition}

\begin{definition}[Equal Opportunity Difference (EOD)~\cite{fair:hardt2016equality}]
Difference in true positive rates across groups:
\[
\text{EOD} = \Pr[\hat{Y} = 1 \mid Y = 1, A = 0] - \Pr[\hat{Y} = 1 \mid Y = 1, A = 1] \text{.}
\]
\end{definition}

\begin{definition}[Statistical Parity Difference (SPD)]
Difference in positive prediction rates across groups:
\[
\text{SPD} = \Pr[\hat{Y} = 1 \mid A = 0] - \Pr[\hat{Y} = 1 \mid A = 1] \text{.}
\]
\end{definition}

All three metrics are defined such that a value of \(0\) corresponds to perfect parity between groups, while larger deviations indicate increasing disparity.

\subsection{Model Training} \label{sub:model_training}

All datasets are split into training (60\%), calibration (20\%), and test (20\%) subsets using a fixed partition. 
Following the standard setting in the privacy-preserving machine learning literature~\cite{yao2025dcr,Ganev2025}, only the training split is treated as the protected dataset. 
In the \dponly{} and \dpfair{} configurations, models are trained on DP synthetic data generated from this real training split, whereas in the \baseline{} and \faironly{} configurations, models are trained directly on the real training split.
In all configurations, final utility and fairness metrics are computed on the real held-out test set. 
The calibration and test splits are disjoint from the protected training database and are never used during DP synthesis. 

The calibration set is only used to tune post-processing fairness mechanisms, while the test set remains unseen during training and calibration, serving exclusively for final evaluation (including post-processing applied at inference). 
Before any fairness post-processing calibration, a model trained on DP synthetic data is a downstream use of the DP output and therefore preserves the DP guarantee with respect to the protected training split. 
After calibration, this guarantee still holds for the original training records used by the synthesizer. 
However, in the main protocol, the calibration records themselves are not protected by this DP guarantee because fairness post-processing methods are calibrated using real held-out data. 
This setup follows a common evaluation protocol in DP synthetic data and privacy attack research, while ensuring consistent comparison across the \baseline{}, \dponly{}, \faironly{}, and \dpfair{} configurations.
To assess whether this calibration protocol affects our conclusions, we further report a DP-compliant calibration ablation in Appendix~\ref{app:ablation_dp_calibration}, where fairness post-processing is calibrated using DP-based calibration records.

\paragraph{Classifier} 
We use a set of three classifiers across all experiments: Extreme Gradient Boosting (\textbf{XGBoost})~\cite{classifier:xgboost}, Logistic Regression (\textbf{LR})~\cite{classifier:LogReg}, and Random Forest (\textbf{RF})~\cite{classifier:RandomForest}. 
These models were selected to cover complementary learning paradigms commonly used for tabular data. 
XGBoost is a widely adopted tree-based ensemble method with strong predictive performance and scalability. 
LR serves as a classical linear baseline, providing a well-understood and interpretable reference point.
RF represents bagging-based ensembles of decision trees, capturing non-linear decision boundaries while remaining robust and widely adopted in practice.
To maintain consistency across configurations, we use the \texttt{binary:logistic} objective and keep all hyperparameters as close to their default values as possible, unless otherwise stated in Appendix \ref{app:classifiers}.

\paragraph{Stability} 
Since DP mechanisms, train/calibration/test splits, synthetic data generation, and classifier training all involve randomness, we repeat each experiment with 20 independent random seeds. 
In the Pareto trade-off plots used in Section~\ref{sec:results}, each point corresponds to the mean across seeds for a fixed configuration, namely dataset, classifier, privacy budget, and fairness intervention.  
To quantify variability across runs, we additionally compute 95\% confidence intervals using Student's \(t\)-distribution. 
Some methods (\eg, LFR in Figure~\ref{fig:results-pareto-aim-f1}) exhibit substantially higher variability across seeds, resulting in intervals that visually dominate the plots; for readability, we omit these intervals from the corresponding figures. 
These statistics are used to assess the robustness and reproducibility of the observed fairness--utility trade-offs.

\subsection{Reproducibility and Extensibility} \label{sub:implementation}

We implement our benchmark on top of the open-source SmartNoise library and the AIF360 fairness toolkit~\cite{fair:AIF360}. 
Our framework integrates DP synthesizers, fairness interventions, and evaluation pipelines in a modular fashion, making it straightforward to add new datasets, generative models, or classifiers. 
More precisely, our benchmark is organized around two main components: 
(1) \emph{DP synthetic data generation}, and 
(2) \emph{execution of experiments}. 
This design provides a clear separation between data generation and downstream evaluation, making the benchmark both reproducible and easily extensible.

\begin{itemize}[leftmargin=*]
    \item \textbf{Synthetic data generation.} 
    This module integrates existing DP synthesizers (AIM, MST) and can be extended with new generators and different data pre-processors. 
    Users can also add additional datasets to the generation pipeline with minimal configuration.

    \item \textbf{Experimental execution.} 
    This module runs end-to-end experiments, including model training, fairness interventions, and metric evaluation, and can be extended by modifying the base classifier. 
\end{itemize}

Together, these two modules ensure that new datasets, synthesizers, classifiers, interventions, or metrics can be integrated with minimal effort.
Moreover, the provided scripts support end-to-end reproduction of the experiments, as well as regeneration of the figures and tables from the corresponding experimental logs.
The full codebase, along with documentation and configuration files, is available in the following GitHub repository \url{https://github.com/vinicius-verona/dp-fair-intervention-benchmark}.
Lastly, to facilitate installation and reuse, the benchmark is also distributed as the open-source Python package BenchmarkDPFair through PyPI at \url{https://pypi.org/project/BenchmarkDPFair/}, and can be installed using \colorbox{gray!12}{\strut\texttt{pip install BenchmarkDPFair.}}

\section{Results and Analysis} \label{sec:results}

We now analyze the empirical behavior of fairness interventions applied to 
DP synthetic data, with a particular focus on \emph{where in the learning pipeline} fairness mechanisms are most effective under privacy constraints.
Throughout this section, we report results obtained with AIM, which is currently regarded as the state-of-the-art DP synthesizer for tabular data, as emphasized in recent benchmarks~\cite{tao2021benchmarking,Rosenblatt2023,qian2023synthcity,Ganev2024} and KDD \& VLDB 2025 tutorials~\cite{Cormode2025}. 
Furthermore, we present the results acquired with the XGBoost algorithm and discuss the other classifiers in Figure~\ref{fig:results-pareto-aim-acc-all-models} and Appendix~\ref{app:add_results}.   
For completeness, additional results using MST are also provided in Appendix~\ref{app:add_results}, with the remaining classifiers; unless otherwise stated, the qualitative trends observed with AIM are consistent across ML models.

Our evaluation spans the four datasets described in Section~\ref{sub:datasets}—three real-world benchmarks (\textbf{Adult}, \textbf{COMPAS}, and \textbf{ACSIncome}) and one synthetic dataset (\textbf{BiasOnDemand})—and compares fairness interventions applied at the 
\emph{pre-processing}, \emph{in-processing}, and \emph{post-processing} stages.
For each dataset, we analyze trade-offs between privacy, utility, and fairness under the four experimental settings of Figure~\ref{fig:methodology}: \baseline{}, \dponly{}, \faironly{}, and \dpfair{}.

\begin{figure*}[!h]
    \centering
    \begin{subfigure}{0.95\textwidth}
        \centering
        \includegraphics[width=\linewidth]{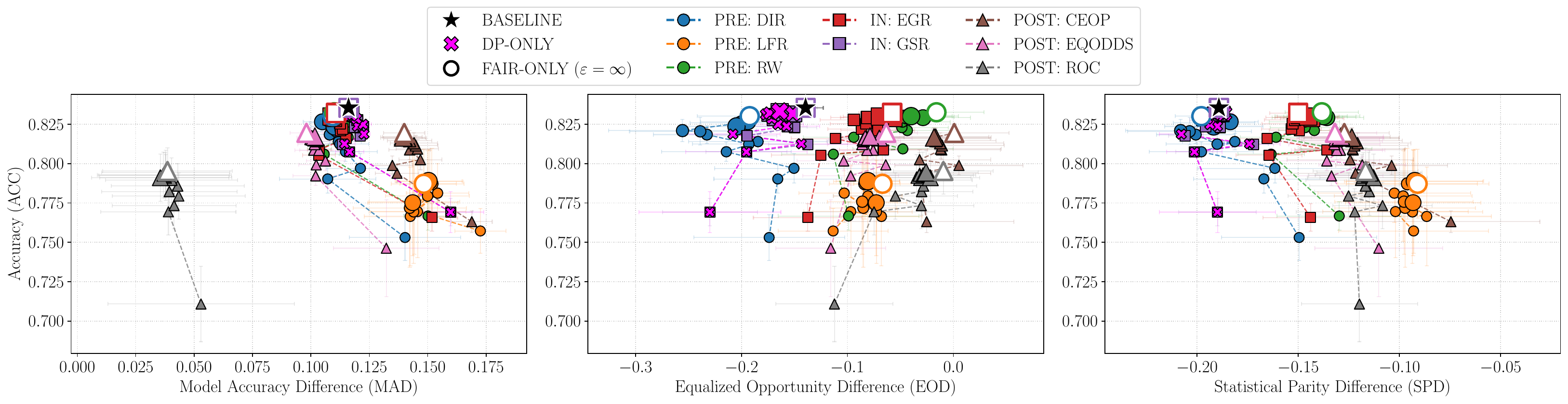}
        \caption{\textbf{Adult}.}
        \label{fig:results-pareto-acc-adult}
    \end{subfigure}\\
    \hfill
    \begin{subfigure}{0.95\textwidth}
        \centering
        \includegraphics[width=\linewidth]{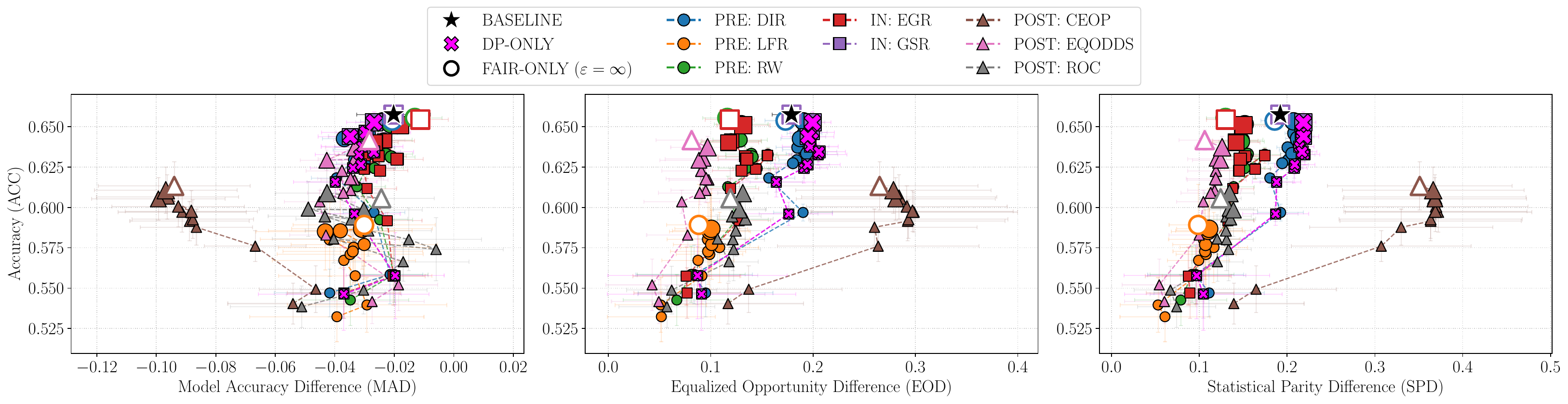}
        \caption{\textbf{COMPAS}.}
        \label{fig:results-pareto-acc-compas}
    \end{subfigure}\\

    \vspace{0.5em}

    \begin{subfigure}{0.95\textwidth}
        \centering
        \includegraphics[width=\linewidth]{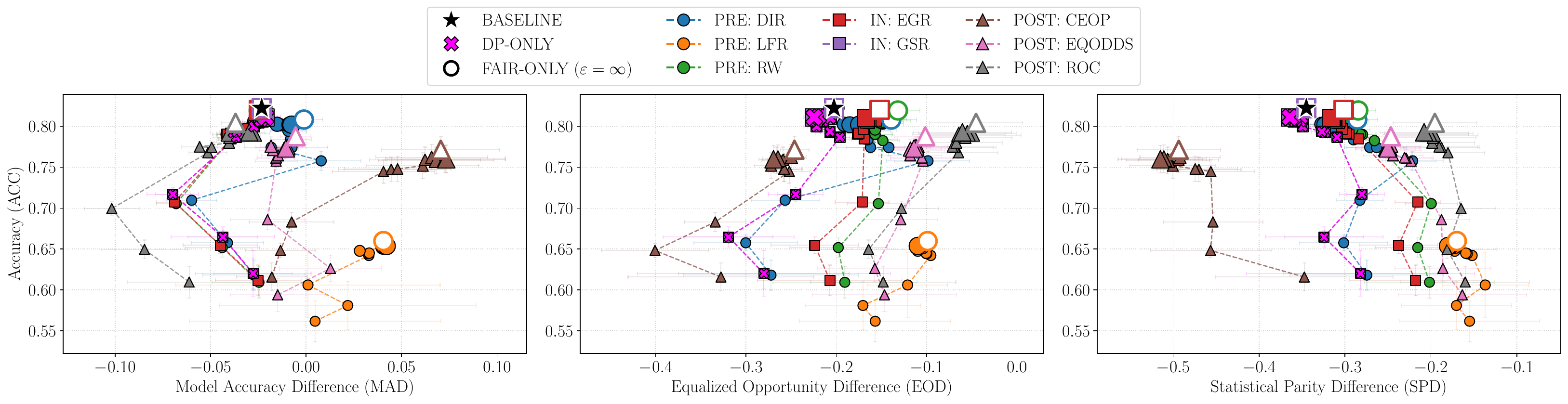}
        \caption{\textbf{ACSIncome}.}
        \label{fig:results-pareto-acc-acsincome}
    \end{subfigure}\\
    \hfill
    \begin{subfigure}{0.95\textwidth}
        \centering
        \includegraphics[width=\linewidth]{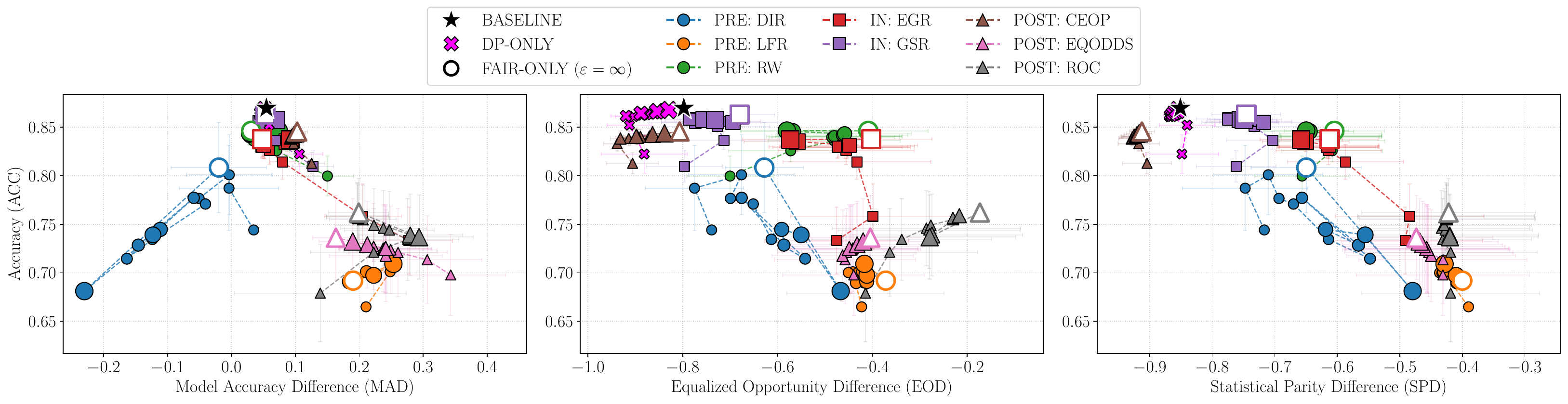}
        \caption{\textbf{BiasOnDemand (Configuration 5)}.}
        \label{fig:results-pareto-acc-bod}
    \end{subfigure}
    \caption{Accuracy--fairness trade-off under the AIM synthesizer across four datasets. Each subfigure reports accuracy (ACC) versus three fairness metrics (MAD, EOD, SPD) for each fairness intervention applied at the pre-, in-, and post-processing stages. 
    Marker size encodes the privacy budget $\varepsilon$ for both \dponly{} and \dpfair{} settings, while hollow markers indicate the \faironly{} setting ($\varepsilon=\infty$).
    The \baseline{} setting (no DP and no fairness intervention) is indicated by the $\star$ symbol.
    }
    \label{fig:results-pareto-aim-acc}
\end{figure*}

\begin{figure*}[!h]
    \centering
    \begin{subfigure}{0.95\textwidth}
        \centering
        \includegraphics[width=\linewidth]{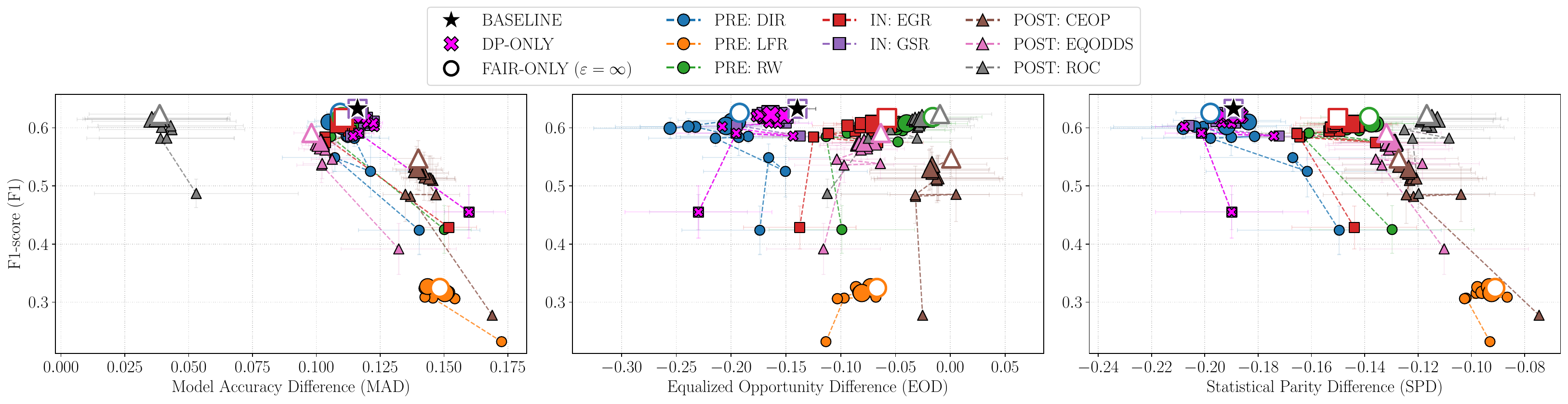}
        \caption{\textbf{Adult}.}
        \label{fig:results-pareto-f1-adult}
    \end{subfigure}\\
    \hfill
    \begin{subfigure}{0.95\textwidth}
        \centering
        \includegraphics[width=\linewidth]{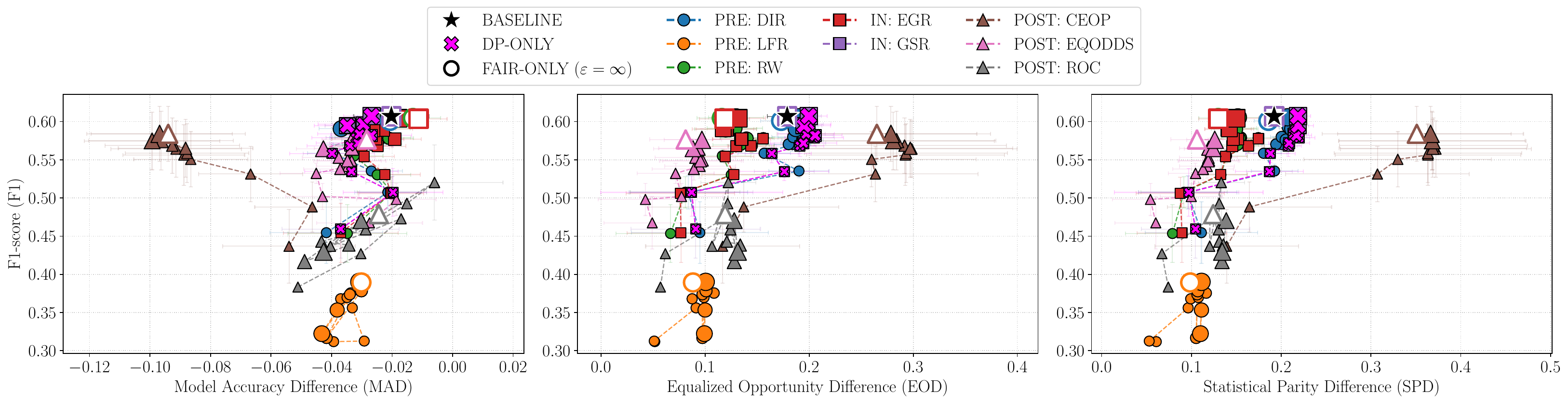}
        \caption{\textbf{COMPAS}.}
        \label{fig:results-pareto-f1-compas}
    \end{subfigure}\\

    \vspace{0.5em}

    \begin{subfigure}{0.95\textwidth}
        \centering
        \includegraphics[width=\linewidth]{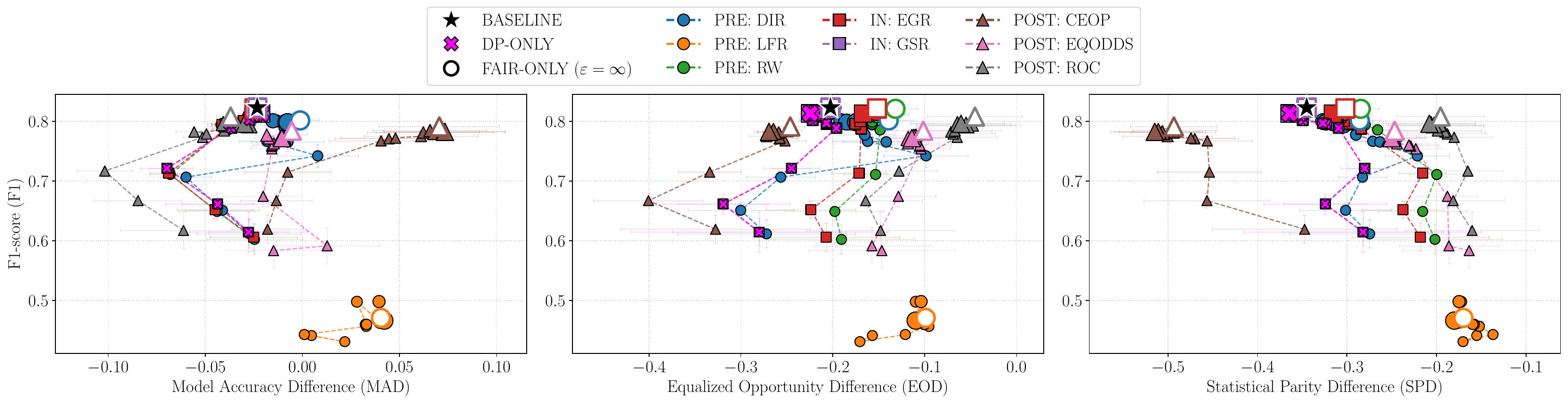}
        \caption{\textbf{ACSIncome}.}
        \label{fig:results-pareto-f1-acsincome}
    \end{subfigure}\\
    \hfill
    \begin{subfigure}{0.95\textwidth}
        \centering
        \includegraphics[width=\linewidth]{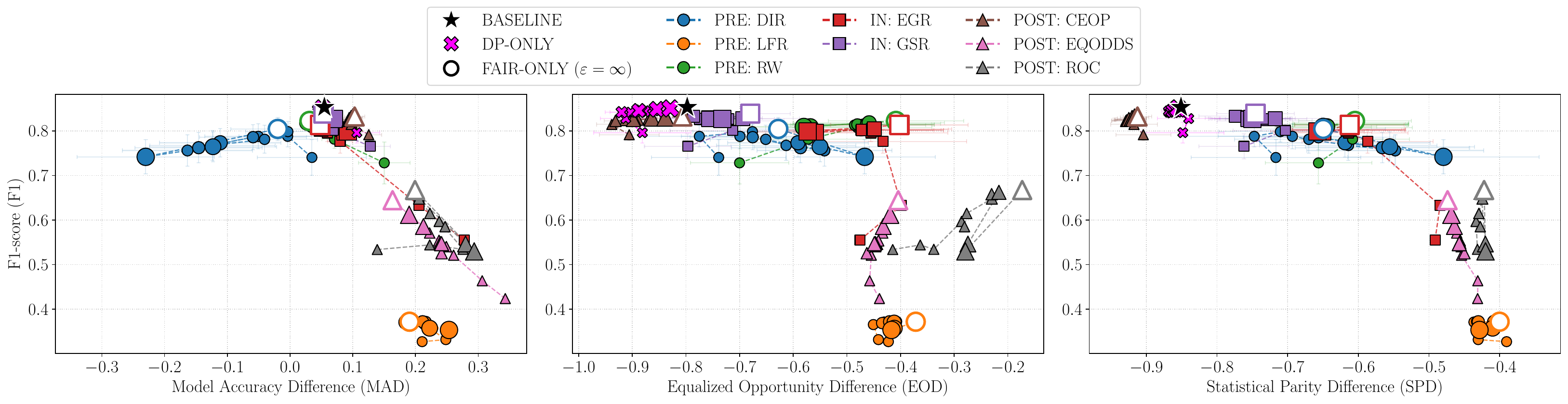}
        \caption{\textbf{BiasOnDemand (Configuration 5)}.}
        \label{fig:results-pareto-f1-bod}
    \end{subfigure}
    \caption{F1-score--fairness trade-offs under the AIM synthesizer across four datasets. 
    Each subfigure reports F1-score versus three fairness metrics (MAD, EOD, SPD) for each fairness intervention applied at the pre-, in-, and post-processing stages. 
    Marker size encodes the privacy budget $\varepsilon$ for both \dponly{} and \dpfair{} settings, while hollow markers indicate the \faironly{} setting ($\varepsilon=\infty$).
    The \baseline{} setting (no DP and no fairness intervention) is indicated by the $\star$ symbol.
    }
    \label{fig:results-pareto-aim-f1}
\end{figure*}

Figures~\ref{fig:results-pareto-aim-acc} and~\ref{fig:results-pareto-aim-f1} present a detailed view of the fairness--utility trade-offs induced by each fairness intervention across all datasets. Specifically, we report:
(i) \textbf{accuracy (ACC)} versus fairness metrics, and (ii) \textbf{F1-score} versus fairness metrics, where F1-score captures the harmonic mean of precision and recall and provides a complementary perspective in the presence of class imbalance.
Each subfigure corresponds to one dataset and plots utility against the three group fairness metrics (MAD, EOD, and SPD). Points correspond to different privacy budgets $\varepsilon$, with marker size encoding the magnitude of $\varepsilon$ (larger markers indicate weaker privacy guarantees), while hollow markers denote the \faironly{} setting ($\varepsilon=\infty$)\footnote{Here, $\varepsilon=\infty$ is used only as a visual shorthand for the non-private fair setting, where fairness interventions are applied to models trained on real data without DP noise; it should not be misperceived with the \baseline{} setting, which is also non-private but does not apply any fairness intervention.}.
This visualization highlights how both utility and fairness evolve as privacy constraints tighten, and how different intervention mechanisms respond to DP noise.

Taken together, these figures provide a method-level view of how fairness interventions interact with DP synthetic data across privacy budgets.
The \faironly{} configuration serves as an upper-bound reference for achievable fairness and utility in the absence of DP noise, while \dponly{} isolates the impact of privacy alone.
The \dpfair{} configurations illustrate the extent to which individual fairness mechanisms can mitigate DP-induced disparities.
In the following subsections, we analyze pre-processing (Section~\ref{sub:results-pre}), in-processing (Section~\ref{sub:results-in}), and post-processing (Section~\ref{sub:results-post}) interventions separately, before discussing our results in Section~\ref{sec:discussion}.

\subsection{Pre-Processing}
\label{sub:results-pre}

We first analyze pre-processing interventions, namely Reweighing (RW)~\cite{fair:Reweighing-paper}, Disparate Impact Remover (DIR)~\cite{fair:DIR}, and Learning Fair Representations (LFR)~\cite{fair:LFR-paper}. These methods operate directly on the training data distribution and therefore constitute the earliest point of intervention in the pipeline.

Across datasets, Figures~\ref{fig:results-pareto-aim-acc} and~\ref{fig:results-pareto-aim-f1} show that some pre-processing methods under the \dpfair{} configuration generally shift predictions away from the \dponly{} region toward the corresponding \faironly{} region in the fairness--utility space. This indicates that pre-processing is capable of correcting not only the bias present in the \baseline{} model, but also a substantial portion of the additional disparities introduced by DP synthetic data. However, these fairness improvements consistently come at the cost of reduced predictive performance, visible as a downward shift in both accuracy and F1-score across privacy budgets.

An additional observation is that pre-processing benefits are only weakly sensitive to the privacy budget $\varepsilon$. While increasing $\varepsilon$ improves the overall utility of DP synthetic data, the relative position of pre-processing methods in the fairness--utility space remains largely stable, suggesting that pre-processing corrections saturate quickly and do not scale strongly with weaker privacy.

\paragraph{Reweighing}
RW reduces group disparities relative to \dponly{}, particularly for SPD, across datasets. In the trade-off plots, RW trajectories shift toward lower disparity while remaining near the \dponly{} utility envelope, indicating moderate fairness gains with limited utility loss. Although RW incurs a systematic decrease in accuracy and F1-score compared to \baseline{} and \faironly{}, this degradation is less pronounced than for other pre-processing methods.
RW therefore provides the most stable and predictable fairness improvements among pre-processing interventions, serving as the strongest baseline in this category. Its gains are most pronounced for SPD, while improvements in error-based metrics such as EOD remain limited, suggesting that RW primarily corrects marginal distributional bias under DP synthetic data.

\paragraph{Disparate Impact Remover}
DIR exhibits limited corrective power under DP synthetic data. In Figures~\ref{fig:results-pareto-aim-acc} and~\ref{fig:results-pareto-aim-f1}, DIR points largely overlap with the \dponly{} region across fairness metrics, indicating minimal reduction in disparity. In some cases, DIR slightly worsens both fairness and utility relative to the \baseline{}. We attribute this behavior to unmet preprocessing assumptions required by DIR~\cite{fair:DIR}, which restrict its ability to decorrelate protected attributes from features. While DIR can be effective when these conditions are satisfied, our benchmark enforces a uniform preprocessing pipeline; under these conditions, DIR remains consistently less effective than RW for DP synthetic data.

\paragraph{Learning Fair Representations}
LFR applies the most aggressive transformation among pre-processing methods. In the Pareto plots of Figures~\ref{fig:results-pareto-aim-acc} and~\ref{fig:results-pareto-aim-f1}, LFR consistently attains the lowest disparity values among pre-processing methods across datasets, often pushing SPD and EOD close to zero for all privacy budgets. This behavior confirms that LFR is effective at enforcing strong group-level parity even under DP synthetic data. However, these fairness gains come at a substantial utility cost: both accuracy and F1-score are systematically reduced, placing LFR in a low-utility region of the trade-off space relative to other interventions.
In practice, this makes LFR less suitable when maintaining predictive performance is a priority, despite its strong parity guarantees.
Beyond this consistent utility degradation, LFR can also exhibit sensitivity to DP noise in certain settings. In particular, for some datasets (notably COMPAS) and specific $\varepsilon$ values or random seeds, the learned representations occasionally collapse one of the target classes, preventing successful downstream model training. These rare but impactful failures are documented in Appendix~\ref{app:add_results}. Overall, while LFR reliably enforces strong fairness, it does so at the expense of both predictive performance and robustness.

\paragraph{Summary}
Overall, pre-processing interventions under \dpfair{} consistently reduce group disparities but do so by trading off predictive performance. RW provides the most stable balance between fairness improvement and utility preservation, DIR remains largely indistinguishable from \dponly{}, and LFR achieves the strongest parity guarantees at the cost of pronounced utility degradation and increased sensitivity to DP noise.

\subsection{In-Processing}
\label{sub:results-in}

We next analyze in-processing interventions, namely Exponentiated Gradient Reduction (EGR) and Grid Search Reduction (GSR)~\cite{fair:Reduction}. Unlike pre-processing methods, in-processing approaches enforce fairness constraints during model training and therefore operate directly on classifiers trained on DP synthetic data.

Across datasets, Figures~\ref{fig:results-pareto-aim-acc} and~\ref{fig:results-pareto-aim-f1} reveal two consistent patterns under the experimental conditions considered in this study. 
First, compared to pre- and post-processing, in-processing interventions achieve more moderate levels of bias correction. 
While both EGR and GSR improve fairness metrics relative to \dponly{}, the resulting reductions in disparity remain bounded and generally do not reach the \faironly{} region of the trade-off space. 
Second, these fairness improvements are obtained while largely preserving predictive performance: accuracy and F1-score under \dpfair{} remain close to the corresponding \dponly{} levels across privacy budgets. 
Taken together, these observations indicate that, under our unified preprocessing pipeline, model configurations, and hyperparameter settings, in-processing methods offer a conservative but stable trade-off, favoring utility preservation while delivering limited yet consistent fairness gains. 
We note that alternative data representations or task-specific tuning could potentially alter this balance; however, such adaptations fall out of the scope of our work, which intentionally enforces comparable conditions across methods.

\paragraph{Exponentiated Gradient Reduction}
Among in-processing approaches, EGR consistently provides the most favorable balance between fairness improvement and utility preservation. 
In the Pareto plots, EGR trajectories move toward lower disparity relative to \dponly{}, particularly for SPD and EOD, while remaining closely aligned with the \dponly{} utility envelope. 
This pattern is observed across all datasets and privacy budgets, indicating that EGR is able to mitigate a non-negligible portion of DP-induced disparities without incurring substantial additional loss in accuracy or F1-score.
However, the extent of fairness correction achieved by EGR remains bounded: while it improves upon \dponly{}, EGR does not fully converge toward the \faironly{} region of the trade-off space. 
Compared to pre- and post-processing methods, its corrective power is more moderate, reflecting a design choice that prioritizes stability and the preservation of utility when training on DP synthetic data.

\paragraph{Grid Search Reduction}
In contrast, GSR exhibits little to no systematic benefit over \dponly{}. 
Across datasets, GSR points largely overlap with the \dponly{} region in the fairness--utility space, indicating negligible disparity reduction. 
While isolated improvements can be observed for specific datasets (\eg, BiasOnDemand in Figures~\ref{fig:results-pareto-acc-bod} and~\ref{fig:results-pareto-f1-bod}), these gains do not generalize and remain inconsistent across privacy budgets. 
Overall, GSR fails to provide reliable fairness improvements under DP synthetic data.

\paragraph{Summary}
Overall, in-processing interventions yield modest and inconsistent fairness gains under DP synthetic data while largely preserving utility. EGR emerges as the only in-processing method that consistently improves fairness relative to \dponly{}, but its impact remains limited compared to pre- and post-processing approaches. 
GSR, by contrast, remains largely indistinguishable from \dponly{} across datasets and privacy regimes. 
These findings suggest that in-processing methods offer a conservative trade-off: they preserve predictive performance, but are insufficient when substantial bias mitigation is required.

\subsection{Post-Processing}
\label{sub:results-post}

We finally analyze post-processing interventions, namely Reject Option Classification (ROC)~\cite{kamiran2012decision}, Equalized Odds Post-Processing (EqOdds)~\cite{fair:hardt2016equality,fair:AIF360}, and Calibrated Equalized Odds (CEOP)~\cite{fair:CEOP}. 
These methods intervene \emph{after} training, adjusting model outputs to satisfy fairness criteria and therefore do not depend on retraining on DP synthetic data.

Across datasets, the method-level Pareto plots (Figures~\ref{fig:results-pareto-aim-acc} and~\ref{fig:results-pareto-aim-f1}) consistently show that post-processing achieves the strongest bias correction \emph{for a given level of utility} under DP synthetic data, particularly for EOD and SPD, and does so more reliably than pre- and in-processing. 
For instance, in the stage-level summaries, POST points systematically move toward lower disparity compared to \dponly{} (most clearly along EOD and SPD), while maintaining accuracy or F1-score within a bounded degradation relative to \dponly{} across privacy budgets.
This behavior highlights a distinctive advantage of post-processing in the DP synthetic setting: because it operates on predictions, it can exploit residual predictive signal that remains after DP synthesis, and translate it into substantial fairness gains without requiring modifications to the (potentially noisy) training distribution.

At the same time, our results indicate that post-processing outcomes are method-dependent. 
ROC and EqOdds consistently shift models toward substantially lower disparities (often approaching the best regions reached by any intervention stage), whereas CEOP exhibits more heterogeneous behavior across datasets and metrics. 
In particular, CEOP can preserve higher utility in some settings, but its fairness correction is less consistent and can introduce metric-specific trade-offs (\eg, improving one notion while leaving another largely unchanged), making it comparatively less reliable as a default choice under DP synthetic data.

\paragraph{Reject Option Classification}
ROC provides the most consistent fairness correction among post-processing methods. 
In the Pareto plots, ROC points typically move toward substantially lower EOD and SPD compared to \dponly{}, frequently placing ROC among the best-performing methods in terms of disparity reduction across datasets. 
This improvement is generally obtained with a moderate utility trade-off: ROC may reduce accuracy and/or F1-score relative to \dponly{}, but remains competitive compared to other intervention stages, and its fairness gains are stable across $\varepsilon$.

\paragraph{Equalized Odds Post-Processing}
EqOdds performs comparably to ROC in terms of fairness correction, consistently moving solutions toward lower disparities across MAD, SPD, and EOD. 
In several datasets, EqOdds provides a favorable balance in the trade-off space, achieving strong reductions in disparity while keeping utility close to \dponly{}. 
Overall, EqOdds is a robust post-processing option whose behavior is stable across privacy regimes.

\paragraph{Calibrated Equalized Odds Post-Processing}
CEOP exhibits the most variable behavior among post-processing methods. 
In the Pareto plots, CEOP can maintain relatively high utility in some settings, but its fairness improvements are less systematic than ROC and EqOdds and can depend on the dataset and the fairness notion considered. 
In particular, CEOP may yield mixed outcomes across metrics (\eg, limited movement on EOD or SPD compared to ROC/EqOdds), and in some cases, its corrections are not aligned across fairness notions. 
This variability suggests that CEOP, at least under the uniform configurations used in our benchmark, is less reliable as a default post-processing choice for DP synthetic data.

\paragraph{Summary}
Overall, post-processing is the most effective and practically reliable intervention stage under DP synthetic data: it consistently achieves the largest reductions in group disparities (especially EOD and SPD) across privacy budgets and datasets. 
ROC and EqOdds are the most robust options, repeatedly delivering strong fairness improvements with bounded utility loss, whereas CEOP is more sensitive to dataset and metric choice and can yield heterogeneous trade-offs. 
These results suggest that \emph{post-processing (particularly ROC and EqOdds) emerges as the most reliable strategy for mitigating bias when training on DP synthetic data, offering the best fairness--utility trade-offs among the evaluated interventions}.

\section{Discussion} \label{sec:discussion}

The results in Section~\ref{sec:results} allow us to revisit the central research question of this work:  \emph{where should one intervene in the ML pipeline to mitigate unfairness when training on DP synthetic data?}

\paragraph{Where to intervene}
Our benchmark reveals that not all points of intervention are equally effective under DP synthetic data.
Pre-processing methods (RW, DIR, LFR) can substantially reduce group disparities, but they typically trade fairness improvements for utility degradation. 
This is expected, as these methods alter the input distribution itself, often at the expense of the signal needed for accurate classification.
In-processing methods (EGR, GSR) offer a more conservative trade-off: they largely preserve utility close to \dponly{}, but achieve only bounded reductions in disparity. 
In particular, EGR yields consistent (yet moderate) improvements, whereas GSR is often indistinguishable from \dponly{}.
Post-processing stands out: ROC and EqOdds consistently achieve the strongest disparity reductions (notably for EOD and SPD) while keeping accuracy and F1-score within a bounded degradation relative to \dponly{} across privacy budgets.
Overall, \emph{\textbf{Overall, the Pareto-front analysis suggests that post-processing is the strongest intervention stage}} in terms of fairness--utility trade-offs.

\paragraph{Mechanism-specific insights}
Several consistent patterns emerge across datasets, privacy budgets, and classifiers when using AIM as the DP synthesizer.
RW reliably reduces group disparities, particularly SPD, but this improvement is typically accompanied by a measurable loss in predictive utility relative to \dponly{}.
LFR enforces the strongest parity guarantees, often driving SPD and EOD close to zero, but does so at a substantial utility cost and exhibits sensitivity to DP noise, occasionally leading to unstable training outcomes.
DIR contributes little under the uniform preprocessing pipeline adopted in our benchmark, with performance largely overlapping the \dponly{} region.
Among in-processing methods, EGR offers a stable compromise: it consistently improves fairness metrics relative to \dponly{} while largely preserving accuracy and F1-score.
However, these gains remain bounded and do not approach the \faironly{} region.
GSR, by contrast, fails to provide reliable or systematic fairness improvements across datasets and privacy regimes.
Post-processing methods are the most robust overall.
ROC and EqOdds consistently appear on or near the Pareto frontier of fairness--utility trade-offs, delivering substantial reductions in EOD and SPD with bounded utility degradation relative to \dponly{}.
CEOP exhibits more heterogeneous behavior, with outcomes that depend on the dataset and fairness metric considered.
Taken together, these results identify \textit{\textbf{ROC and EqOdds as the most reliable mechanisms for balancing privacy, fairness, and utility under DP synthetic data generated by AIM}}.

\paragraph{Why post-processing outperforms other interventions}
Post-processing mechanisms tend to outperform other interventions under DP synthetic data \emph{in our benchmark} because they act at the decision stage, directly on model predictions rather than on noisy training data.
Methods such as ROC and EqOdds adjust decision thresholds or predicted probabilities, making them comparatively robust to distortions introduced by DP synthesis.
Unlike pre- and in-processing approaches that must learn fairness corrections from perturbed features and labels, post-processing leverages residual separability in the model's output space to realign group outcomes with bounded changes in accuracy and F1-score.
This mechanism-level explanation is consistent with the empirical patterns observed in the Pareto-front analysis.

\paragraph{Synthesizer and classifier effects}
The relative effectiveness of fairness interventions depends on both the DP synthesizer and, to a lesser extent, the classifier. In the main paper, we focus on AIM, which retains sufficient predictive utility for fairness interventions to operate meaningfully. Under AIM, the qualitative stage-level trends remain stable across the three classifiers considered, as shown in Figure~\ref{fig:results-pareto-aim-acc-all-models} and further detailed in Appendix~\ref{app:add_results}. Although absolute utility and disparity values vary across model families, pre-, in-, and post-processing interventions occupy comparable regions of the fairness--utility space. LR exhibits larger sensitivity on the Adult dataset, particularly for recall-sensitive fairness metrics, which is consistent with its linear hypothesis class and different baseline error structure compared to tree-based models (\ie, XGBoost and RF). Importantly, these quantitative differences do not overturn the overall stage-level conclusion.

At the same time, the achievable fairness--utility trade-offs depend on the synthetic data generator. With AIM, several interventions move \dpfair{} outcomes toward their \faironly{} counterparts, although the gap to \faironly{} is not always fully closed and depends on the intervention stage and fairness metric. In contrast, results with MST reveal a different limitation: while fairness interventions under \dpfair{} can reduce disparities, the corresponding utility levels remain substantially below those achieved by \faironly{}, even as the privacy budget increases. This persistent gap suggests that, under MST, DP synthetic data imposes a structural ceiling on achievable utility that fairness interventions cannot overcome. As a result, improvements in fairness must be interpreted relative to a constrained utility regime, rather than as steps toward the non-private fair optimum. Practitioners should therefore consider not only \emph{where} to intervene, but also whether the chosen DP synthesizer preserves sufficient utility headroom for fairness interventions to approach non-private performance.

\begin{figure*}[!h]
    \centering
    \includegraphics[width=\linewidth]{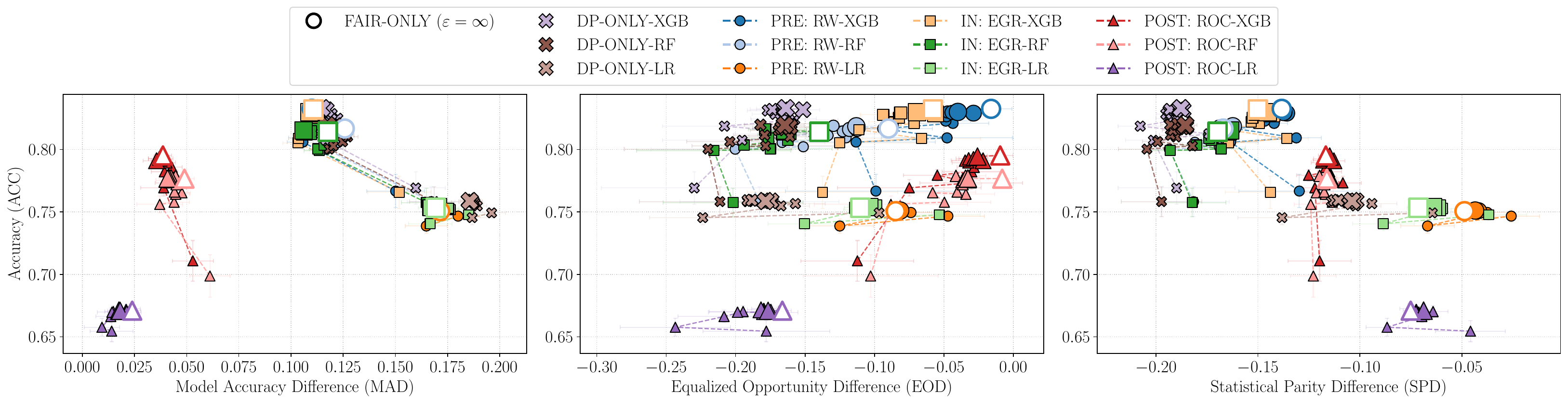}
    \caption{Accuracy--fairness trade-offs under the AIM synthesizer across the Adult dataset and all tested ML models (XGBoost, Random Forest, and Logistic Regression). 
    Each subfigure reports accuracy across three fairness metrics (MAD, EOD, SPD) for representative interventions from each stage. 
    Marker size encodes the privacy budget $\varepsilon$ for both \dponly{} and \dpfair{} settings, while hollow markers indicate the \faironly{} setting ($\varepsilon=\infty$).
    }
    \label{fig:results-pareto-aim-acc-all-models}
\end{figure*}

\paragraph{Ablations and robustness checks}
We further assess whether two methodological choices affect our conclusions.
First, to verify that the advantage of post-processing is not driven by access to a real held-out calibration set, we repeat the post-processing experiments using DP-compliant calibration data.
As reported in Appendix~\ref{app:ablation_dp_calibration}, the qualitative conclusions remain unchanged: ROC and EqOdds continue to provide the strongest fairness--utility trade-offs.
Second, to examine whether the more limited gains of in-processing methods are due to default hyperparameter choices, we conduct a hyperparameter-sensitivity analysis for EGR and GSR.
The results, reported in Appendix~\ref{app:ablation_in_processing}, show that tuning can improve some local trade-offs but does not alter the overall stage-level conclusion: in-processing remains more conservative than post-processing under DP synthetic data.

\paragraph{Implications}
The broader implication of our findings is that fairness interventions are not uniformly transferable from non-private to DP settings. 
While post-processing remains reliable, pre- and in-processing methods are more fragile under DP-induced distributional shifts. 
Moreover, the utility--fairness trade-off interacts strongly with the synthesizer: mechanisms that are effective with AIM may have a limited effect under MST. 
For practitioners, this suggests two guidelines.
First, when utility preservation is a priority, combining a high-utility DP synthesizer such as AIM with post-processing interventions (in particular ROC or EqOdds) yields the most favorable trade-offs.
Second, when enforcing stronger parity constraints is critical, and some utility loss is acceptable, pre- or in-processing methods such as RW or EGR may be considered, provided their costs and stability limitations are carefully evaluated.

\paragraph{Key takeaway and statistical validation.}
Overall, our benchmark shows that although DP amplifies fairness--utility trade-offs when training on synthetic data, carefully chosen intervention strategies can still recover part of the fairness loss while preserving predictive utility. 
To complement the Pareto-front analysis, we statistically validate two central claims using one-sided paired Wilcoxon signed-rank tests~\cite{conover1999practical} across the four main datasets, three classifiers (XGBoost, RF, and LR), and AIM as the DP synthesizer.

\begin{itemize}[leftmargin=*]
    \item \textbf{Claim 1: \dpfair{} recovers fairness degradation caused by \dponly{}.}
    For each fairness metric \(m \in \{\mathrm{MAD}, \mathrm{EOD}, \mathrm{SPD}\}\), we pair configurations with the same dataset, classifier, privacy budget \(\varepsilon\), seed, and DP synthesizer, and compute
    \[
    d_i = |m^{(i)}_{\mathrm{DP+Fair}}| - |m^{(i)}_{\mathrm{DP-only}}|.
    \]
    We test the one-sided alternative that \(d_i<0\), where negative values indicate that the fairness intervention reduces the absolute disparity observed under \dponly{}. 
    As shown in Table~\ref{tab:claim1-wilcoxon-global}, \dpfair{} significantly reduces EOD and SPD, with confidence intervals entirely below zero, while MAD does not improve globally. 
    Thus, Claim 1 is supported for opportunity- and parity-based group fairness metrics, while the MAD result shows that fairness recovery under \dpfair{} is metric-dependent.
    
    \setlength{\tabcolsep}{4pt}
    \begin{table}[!htb]
        \centering
        \begin{tabular}{cccccc}
        \toprule
        Metric &    $n$ & $\bar{d}$ &            95\% CI &    $p$-value & Win rate \\
        \midrule
           MAD & 20,998 &    0.0204 &   [0.0194, 0.0213] &        1.000 &   40.0\% \\
           EOD & 20,998 &   -0.1185 & [-0.1208, -0.1162] & $<10^{-16}$ &   72.6\% \\
           SPD & 20,998 &   -0.0879 & [-0.0898, -0.0859] & $<10^{-16}$ &   74.4\% \\
        \bottomrule
        \end{tabular}
        \caption{Paired Wilcoxon signed-rank analysis for Claim 1 under AIM across the four main datasets and three classifiers. Negative $\bar{d}$ indicates that \dpfair{} reduces the absolute fairness gap relative to \dponly{}.}
        \label{tab:claim1-wilcoxon-global}
    \end{table}

    \item \textbf{Claim 2: POST provides stronger EOD/SPD-oriented fairness--utility trade-offs than PRE/IN.}
    To jointly compare utility and fairness, we use a weighted Euclidean distance to the ideal point \((U=1, |m|=0)\). For each dataset, classifier, seed, privacy budget \(\varepsilon\), utility metric, and fairness metric, we compute
    \begin{equation}  \label{eq:objective_weighted_euclidean_dist}
    S_a^{(i)}
    =
    \sqrt{
    w_U\left(1-U_a^{(i)}\right)^2
    +
    w_F\left|m_a^{(i)}\right|^2
    },
    \end{equation}
    \noindent where \(U\in\{\mathrm{ACC},\mathrm{F1}\}\), \(m\in\{\mathrm{EOD},\mathrm{SPD}\}\), and \(w_U=w_F=0.5\). The square root reflects the Euclidean-distance interpretation; lower values indicate better fairness--utility trade-offs. For each intervention stage \(s\), we define $S_s^{(i)}=\min_{a\in s} S_a^{(i)}$, and compare POST against PRE and IN through
    \[
    d_i
    =
    S_{\mathrm{POST}}^{(i)}
    -
    S_{\mathrm{OTHER}}^{(i)},
    \qquad
    \mathrm{OTHER}\in\{\mathrm{PRE},\mathrm{IN}\}.
    \]
    We again test the one-sided alternative that \(d_i<0\). Negative values indicate that POST is closer to the ideal fairness--utility point. As shown in Table~\ref{tab:claim2-scalarized-wilcoxon-all-models}, POST, defined as the best score among ROC and EqOdds, significantly outperforms both PRE and IN for EOD/SPD-oriented trade-offs.

    \setlength{\tabcolsep}{2.5pt}
    \begin{table}[!htb]
    \centering
    \begin{tabular}{cccccc}
    \toprule
     Comparison &   $n$ & $\bar{d}$ &            95\% CI &    $p$-value & Win rate  \\
    \hline
    POST vs PRE & 11,088 &   -0.0199 & [-0.0207, -0.0190] & $<10^{-16}$ &   72.8\% \\
     POST vs IN & 11,088 &   -0.0294 & [-0.0303, -0.0286] & $<10^{-16}$ &   79.2\%  \\
    \bottomrule
    \end{tabular}
    \caption{Paired Wilcoxon signed-rank analysis for Claim~2 under AIM across the four main datasets and three classifiers. 
    The analysis is restricted to EOD/SPD-oriented trade-offs, with POST defined as the best score among ROC and EqOdds. Negative $\bar{d}$ indicates that POST achieves a lower scalarized fairness--utility score than the compared stage.}
    \label{tab:claim2-scalarized-wilcoxon-all-models}
    \end{table}

\end{itemize}

These tests support the main EOD/SPD-oriented conclusions while also showing that the findings remain criterion-dependent, since MAD does not improve globally, and POST superiority is established under the scalarized EOD/SPD trade-off in Equation~\eqref{eq:objective_weighted_euclidean_dist}.

\section{Conclusion, Limitations, and Perspectives} \label{sec:conclusion}

\paragraph{Concluding remarks}
We introduced, to our knowledge, the first systematic benchmark of fairness-aware learning on \emph{DP synthetic tabular data}, covering multiple fairness intervention stages (pre-, in-, and post-processing), four datasets, a wide range of privacy budgets, and three different ML classifiers. 
Our main analysis centers on AIM~\cite{dp:aim}, a state-of-the-art marginal-based DP synthesizer for tabular data, with additional results for MST~\cite{dp:mst} reported in Appendix~\ref{app:add_results}.
Across configurations, we find that while DP alone generally reduces utility and can exacerbate group disparities, \emph{fairness interventions can partially recover fairness under DP synthetic data}.
Within the evaluated scope of AIF360-based group fairness mechanisms and marginal-based DP synthesizers, \emph{post-processing methods tend to provide the most stable fairness--utility trade-offs under AIM}, particularly at moderate privacy budgets.
In particular, methods such as ROC~\cite{kamiran2012decision} and EqOdds~\cite{fair:hardt2016equality} frequently reduce group disparities while incurring bounded utility loss relative to \dponly{} training.
These results provide \emph{actionable guidance} on where and how to intervene in DP-synthetic learning pipelines, while highlighting that both the intervention stage and the DP synthesizer shape the attainable privacy--fairness--utility trade-off.

\paragraph{Limitations}
Our study focuses on \emph{tabular, binary classification} with a \emph{binary protected attribute}. 
While this represents the most widely studied and standardized setting in the fairness literature~\cite{Barocas2023,fair:review,fair-dp:fioretto2022,yao2025sok},
extensions to multi-class tasks, regression problems, and multi-valued or intersectional protected attributes are important to cover a broader range of real-world scenarios. 
Such extensions, however, would require substantial adaptations of fairness metrics, mitigation strategies (which currently rely on the AIF360 framework and its binary attribute/label formulations), and evaluation protocols, and are therefore left for future work.
Moreover, we restrict our evaluation to \emph{marginal-based} DP synthetic data generators, as prior benchmarks~\cite{tao2021benchmarking,Pereira2024,Ganev2024} show that they consistently outperform \emph{deep generative} DP synthesizers on tabular data under realistic privacy budgets. As a result, our conclusions may not directly transfer to settings where lower-utility or domain-specific DP generative models are employed.
Regarding learning algorithms, we evaluate \emph{three representative base classifiers} (XGBoost, RF, and LR) using default
hyperparameters, unless otherwise stated, to ensure controlled and reproducible comparisons. 
Other model families or extensively tuned configurations may interact differently with both DP noise and fairness interventions.
Finally, our fairness evaluation focuses on \emph{group-level accuracy and error disparity metrics}. 
Broader notions of fairness, such as full Equalized Odds, counterfactual fairness, or individual fairness, are not considered and remain important directions for extending this study.

\paragraph{Future directions}
Building on this benchmark, several extensions remain open.
These include evaluating \emph{additional families of DP synthesizers} (\eg, DP-GANs~\cite{torkzadehmahani2019dp}), extending to \emph{multi-task, multi-label, and multi-class settings} with richer and potentially intersecting protected attributes, and assessing a wider range of \emph{learning algorithms} and \emph{hyperparameter regimes}.
Future work should also expand beyond AIF360-based interventions by considering newer fairness mechanisms, broaden the metric suite to include \emph{calibration-based}, \emph{individual}, or \emph{counterfactual fairness} criteria, and compare post-hoc mitigation strategies with approaches that jointly enforce fairness and privacy during synthetic data generation (\eg,~\cite{PFGuard}).
To facilitate progress along these directions, we release all code and experimental artifacts (see Section~\ref{sub:implementation}).


\begin{acks}
\noindent This work was partially supported by the French National Research Agency (ANR) research grants (ANR-24-CE23-6239, ANR-23-IACL-0006) and by the Natural Sciences and Engineering Research Council of Canada (NSERC) Discovery grant (RGPIN-2026-04945).
\end{acks}

\section*{AI-Generated Content Acknowledgement}
\noindent The authors acknowledge the use of ChatGPT (OpenAI, GPT-5 model) and Claude (Anthropic, Sonnet-4.6 model) to assist with language-related improvements, including grammar correction, spelling, formatting consistency, and refinement of phrasing for clarity and readability. 
Moreover, the authors acknowledge the use of the aforementioned models as auxiliaries to build automation shell scripts to facilitate artifact generation.
All conceptual, technical, and experimental contributions originate solely from the authors.

\bibliographystyle{ACM-Reference-Format}
\bibliography{references}


% \clearpage
% \newpage
% \onecolumn
\appendix

\section{Classifiers, Datasets, \& Data Pre-Processing} \label{app:datasets}

This section provides complementary details to support the reproducibility of the experimental results presented in the main paper. We first describe the data pre-processing pipeline applied to each dataset, followed by additional results for alternative classifier and synthesizer configurations. The data treatment pipeline is structured into three stages:
\begin{enumerate*}[label=(\roman*)]
    \item data cleaning,
    \item feature encoding, and
    \item binary transformation.
\end{enumerate*}
For some datasets, feature domain compression is additionally applied, following the recommendations of~\cite{dp:aim}. When used, this compression is explicitly described under \textbf{Feature encoding} for the corresponding dataset.

\subsection{Adult Dataset}
    \textbf{Data Cleaning.} First, we start by removing all samples with one or more missing features (null samples) and, consequently, their label. Then, we proceeded to relabel the dataset so that only two labels remained ($\leq 50K$ and $> 50K$). Lastly, we remove the following features from the dataset: 
    \begin{enumerate*}[label=(\roman*)]
        \item \textit{fnlwgt};
        \item \textit{education-num}; 
        \item \textit{capital-loss}; 
        \item \textit{capital-gain}.
    \end{enumerate*}
    By the end of this step, the remaining dataset contains 10 features and $47,621$ samples. 

    \textbf{Feature encoding.} The first step is to compress the domain of the dataset (as cardinality would pose a problem for AIM or MST). Such a task was performed by grouping the \textit{ages} into 20 different bins, each with a 5-year range (1-5, 6-10, and so on). Then, we perform the same for the \textit{hours-per-week} feature, using five different sets with a 20-hour range. Lastly, we compress the \textit{native-country} feature by directly mapping each unique country to a continent. The next step in the encoding pipeline is to ensure all non-numeric features are transformed into numerical features. This is done using pandas categorical encoding, which assigns an integer to each unique value per feature.

    \textbf{Binary transformation.} The last step of the pipeline is the binary transformation of each sensitive feature. This task is performed once for all sensitive features to facilitate tests with different values for the protected attribute \(A\). For the Adult dataset, this transformation is as follows:
    \begin{enumerate*}[label=(\roman*)]
        \item maps \say{Male} \textit{sex} to 1 and others to 0;
        \item maps \say{White} \textit{race} to 1 and others to 0.
    \end{enumerate*}
    
\subsection{COMPAS Dataset}
    \textbf{Data Cleaning.} First, we start by removing all samples where the \textit{race} is neither \textbf{African-American} nor \textbf{Caucasian}. Then we remove all columns except for 
     \begin{enumerate*}[label=(\roman*)]
        \item \textit{sex},
        \item \textit{race}, 
        \item \textit{age},
        \item \textit{c\_charge\_degree},
        \item \textit{prior\_counts},
        \item \textit{score\_text}, and
        \item \textit{two\_year\_recid} (target column).
    \end{enumerate*}
    Then, as in Adult, we remove all samples with one or more missing features (null samples) and, consequently, their labels.
    By the end of this step, the remaining dataset contains 7 features and $5,050$ samples. 

    \textbf{Feature encoding.} The first step is to ensure all non-numeric features are transformed into numerical features. This is done exactly like in the Adult dataset, using pandas categorical encoding.

    \textbf{Binary transformation.} The last step of the pipeline is the binary transformation of each sensitive feature. For the COMPAS dataset, this transformation is as follows:
    \begin{enumerate*}[label=(\roman*)]
        \item maps \say{Male} \textit{sex} to 1 and others to 0;
        \item maps \say{Caucasian} \textit{race} to 1 and others to 0.
        \item maps \textit{age} $\geq 25$ and \textit{age} $\leq 45$ to 1 and others to 0.
    \end{enumerate*}
\subsection{ACSIncome Dataset}
    \textbf{Data Cleaning.} First, we start by filtering the census to the year 2018 and the state of Utah. Then we remove all features except for: 
     \begin{enumerate*}[label=(\roman*)]
        \item \textit{AGEP},
        \item \textit{COW}, 
        \item \textit{SCHL},
        \item \textit{MAR},
        \item \textit{OCCP},
        \item \textit{POBP},
        \item \textit{RELP},
        \item \textit{WKHP},
        \item \textit{SEX},
        \item \textit{RAC1P}, and
        \item \textit{PINCP} (target column).
    \end{enumerate*}
    Then, as in Adult and COMPAS, we remove all samples with one or more missing features  (null samples) and, consequently, their labels.
    By the end of this step, the remaining dataset contains 11 features and $16,337$ samples. 

    \textbf{Feature encoding.} Like in Adult, the first step is to compress the domain of large features, such as \textit{OCCP}. This is done by grouping each occupation into 6 categories, following the code list \cite{USCensus_ACSPUMS2023CodeLists}. We then ensure that all non-numeric features are transformed into numerical features. This is done exactly like in the Adult dataset, using pandas categorical encoding.

    \textbf{Binary transformation.} The last step of the pipeline is the binary transformation of each sensitive feature. For the ACSIncome dataset, this transformation is as follows:
    \begin{enumerate*}[label=(\roman*)]
        \item maps \say{Male} \textit{sex} to 1 and others to 0;
        \item maps \say{White alone} \textit{race} to 1 and others to 0.
    \end{enumerate*}
    
\subsection{BiasOnDemand Dataset}
    BiasOnDemands (BoD) has a simpler pipeline, as it is a fully customised dataset. Using the BoD dataset generator with the configuration described in Table~\ref{tab:BoD}, we can skip the data cleaning step and proceed to feature encoding\footnote{A description for each parameter can be found here: \url{https://github.com/rcrupiISP/BiasOnDemand/tree/main}}.

    \textbf{Feature encoding.} This step first discretises all features in the dataset that have continuous values (\(R,Q\)). For that, the values are grouped into 5 different bins with integer values representing each bin.

    \textbf{Binary transformation.} Essentially, all that needs to be done is to invert the values of the protected attribute \(A\), to follow the pattern of the other datasets.

    \begin{table*}[!htb]
    \centering
    \rowcolors{2}{gray!15}{white} 
    \begin{tabular}{l|c|c|c|c|c|c}
    \toprule
    \textbf{Parameter} & \textbf{Config 1} & \textbf{Config 2} & \textbf{Config 3} & \textbf{Config 4} & \textbf{Config 5} & \textbf{Config 6} \\
    \midrule
    $l_y$ &  10 &  0 &  0 & 0 & 0 & 0\\
    $l_{my}$ &  0 &  15 &  0 & 0 & 0 & 0\\
    $thr\_supp$ &  1 &  1 &  1 & 1 & 1 & 1\\
    $l_{hr}$ &  0 &  0 &  0 & 10 & 0 & 10\\
    $l_{hq}$ &  0 &  0 &  0 & 0 & 10 & 0\\
    $l_m$ &  0 &  0 &  0 & 0 & 0 & 0\\
    $p_u$ &  1 &  1 &  0.2 & 1 & 1 & 1\\
    $l_r$ &  false &  false &  false & false & false & false\\
    $l_o$ &  false &  false &  false & false & false & false\\
    $l_{yb}$ &  0 &  0 &  0 & 0 & 0 & 0\\
    $l_q$ &  2 &  2 &  2 & 2 & 10 & 2\\
    $sy$ &  5 &  5 &  5 & 5 & 5 & 5\\
    $l_{rq}$ &  0 &  0 &  0 & 0 & 0 & 10\\
    $l_{my}\_non\_linear$ &  false &  false &  false & false & false & false\\
    \bottomrule
    \end{tabular}
    \caption{Configuration parameters for BoD data generation.}
    \label{tab:BoD}
    \end{table*}

\subsection{Classifiers} \label{app:classifiers}
In this section, we report any modifications to the default values of the hyperparameters of each classifier, and thereafter, the motivation for each parameter. Both XGBoost and Random Forest are executed in their default settings, with only the random seed being set, for reproducibility. As for the Logistic Regression, we refer to Table \ref{tab:LogRegParams} to display the parameters. The motivation behind this set of parameters is mostly empirical experimentation. 
Prior results with logistic regression under default parameters did not reach reasonable utility metrics (compared to the other two), either due to high-dimensionality and correlation of the data, over/under-fitting, or the non-linear structure of the data. Under these conclusions, we experimented with different sets of parameters (via Grid Search optimization) until reaching reasonable utility metrics. The parameters in Table \ref{tab:LogRegParams} were chosen as an attempt to balance L1 and L2 regularization while improving the reliability of the convergence in high-dimensional and correlated feature spaces. 

\begin{table}[!htb]
    \centering
    \rowcolors{2}{gray!15}{white} 
    \begin{tabular}{l|c}
    \toprule
    \textbf{Parameter} & \textbf{Choice} \\
    \midrule
    solver & saga\\
    penalty & elasticnet\\
    l1\_ratio & 0.5\\
    C & 0.8\\
    max\_iter & 10000\\
    \bottomrule
    \end{tabular}
    \caption{Configuration parameters for Logistic Regression classifier.}
    \label{tab:LogRegParams}
    \end{table}

\section{Ablation and Additional Results} \label{app:ablation_add_results}

This section provides complementary analyses that support the main findings reported in Section~\ref{sec:results}. 
We first present two ablation studies targeting key methodological choices: (i) whether the advantage of post-processing methods depends on access to a real held-out calibration set, and (ii) whether the comparatively limited gains of in-processing methods are due to their default hyperparameter settings. 
We then report additional experimental results across alternative classifier and synthesizer configurations, as well as extended BiasOnDemand settings. 
Together, these analyses help assess whether the observed fairness--utility--privacy trade-offs are stable across calibration protocols, in-processing tuning choices, model families, DP synthesizers, and controlled bias scenarios.

\subsection{Ablation: DP-Compliant Calibration Set for Post-Processing} \label{app:ablation_dp_calibration}

In the main benchmark (Section~\ref{sub:overview_benchmark}), fairness post-processing methods are calibrated using a real held-out calibration set. 
As discussed in Section~\ref{sub:model_training}, this does not affect the DP guarantee with respect to the protected training split used by the synthesizer, but the calibration records themselves are not protected by that guarantee. 
To assess whether this design choice biases the comparison in favor of post-processing methods, we conduct an ablation in which ROC, EqOdds, and CEOP are calibrated using DP-compliant calibration data rather than non-private calibration records.

In contrast to the methodology described in Section \ref{sub:model_training}, the partitioning of training, calibration, and test subsets is performed as follows:
\begin{enumerate*}
    \item The initial dataset $D$ is partitioned into disjoint subsets ${D}_{\text{temp}}$ and ${D}_{\text{test}}$ using an 80-20 split, where $|{D}_{\text{test}}| = 0.2 \times |{D}|$.
    \item The temporary set ${D}_{\text{temp}}$ undergoes differential privacy (DP) preprocessing to ensure $\epsilon$-DP compliance.
    \item Following this processing, ${D}_{\text{temp}}$ is further partitioned into ${D}_{\text{train}}$ and ${D}_{\text{cal}}$ with a 75-25 split ratio, yielding $|{D}_{\text{train}}| = 0.75 \times |{D}_{\text{temp}}| = 0.6 \times |{D}|$ and $|{D}_{\text{cal}}| = 0.25 \times |{D}_{\text{temp}}| = 0.2 \times |{D}|$.
\end{enumerate*}

This partitioning yields three final subsets with the following proportions relative to the original dataset: DP-processed training set (60\%), DP-processed calibration set (20\%), and untouched test set (20\%), as in Section \ref{sub:model_training} proportions. We note that for experimental configurations where post-processing calibration is not required (\ie, pre-processing or in-processing fairness interventions), ${D}_{\text{cal}}$ is excluded from the pipeline. For the \baseline{} and \faironly{} classifiers, all three subsets remain in their original, untouched form, maintaining the same distributional partitioning scheme as described above.

\begin{figure*}[!ht]
    \centering
    \begin{subfigure}{0.99\textwidth}
        \centering
        \includegraphics[width=\linewidth]{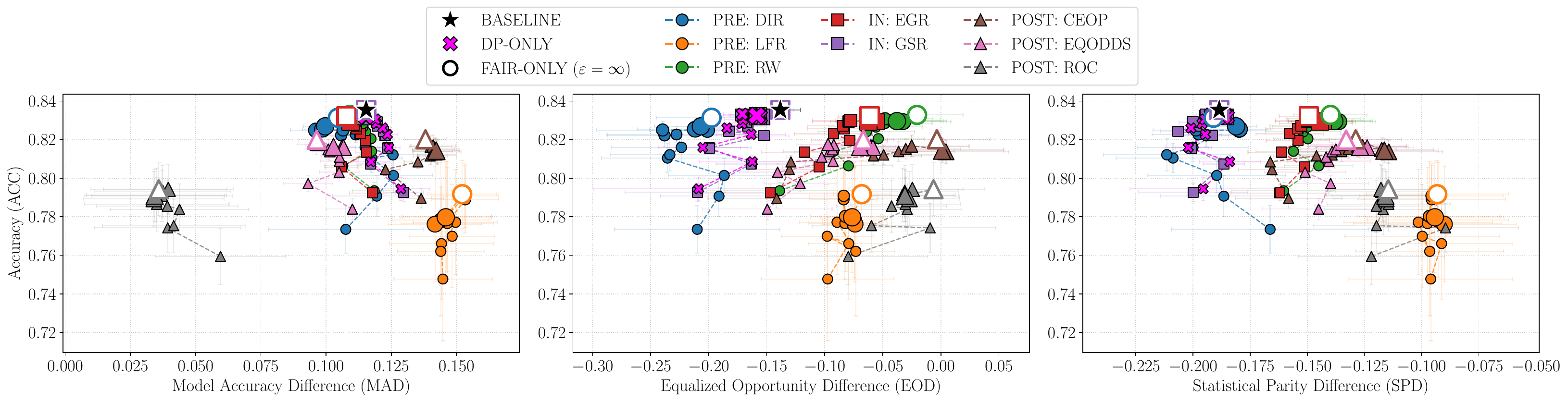}
        \caption{\textbf{Adult}.}
        \label{fig-app:results-pareto-acc-adult-xgb-ablation}
    \end{subfigure}\\
    \hfill
    \begin{subfigure}{0.99\textwidth}
        \centering
        \includegraphics[width=\linewidth]{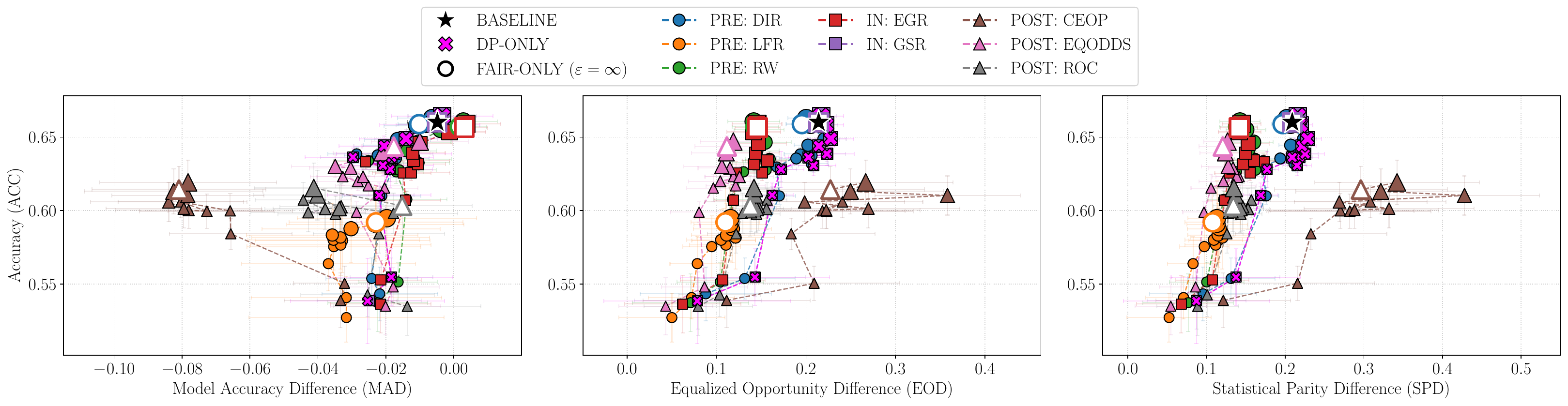}
        \caption{\textbf{Compas}.}
        \label{fig-app:results-pareto-acc-compas-xgb-ablation}
    \end{subfigure}\\
    \hfill
    \begin{subfigure}{0.99\textwidth}
        \centering
        \includegraphics[width=\linewidth]{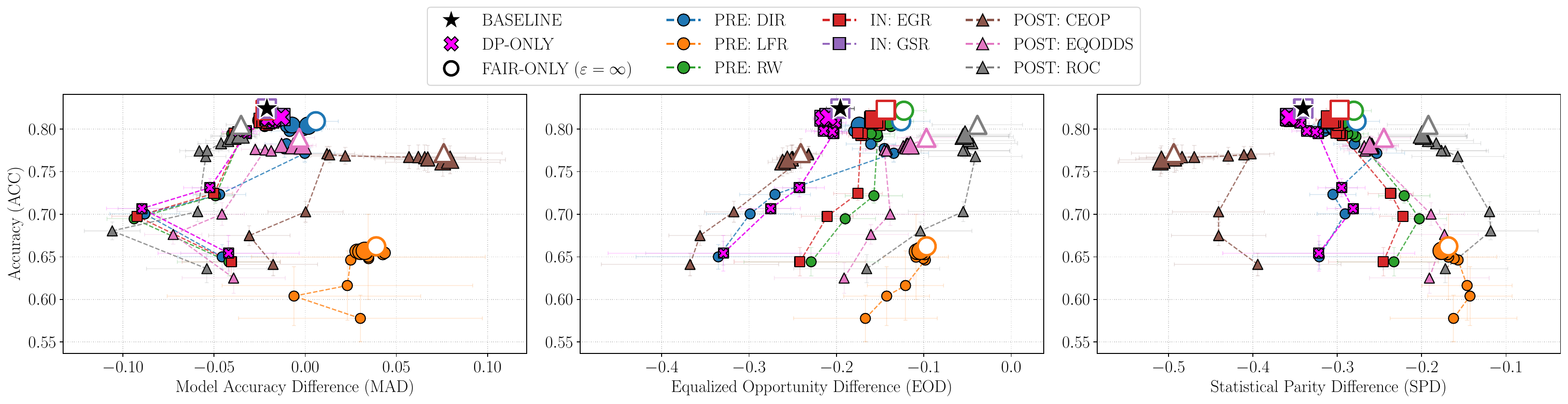}
        \caption{\textbf{ACSIncome}.}
        \label{fig-app:results-pareto-acc-acsincome-xgb-ablation}
    \end{subfigure}
    \caption{Accuracy--fairness trade-off under the \textbf{AIM} synthesizer across three real-world datasets (Adult, Compas and ACSIncome) with the \textbf{XGBoost} classifier. Each subfigure reports accuracy (ACC) versus three fairness metrics (MAD, EOD, SPD) for each fairness intervention applied at the pre-, in-, and post-processing stages. The post-processing \dpfair{} settings are calibrated using a DP synthetic calibration set. 
    Marker size encodes the privacy budget $\varepsilon$ for both \dponly{} and \dpfair{} settings, while hollow markers indicate the \faironly{} setting ($\varepsilon=\infty$).
    The \baseline{} setting (no DP and no fairness intervention) is indicated by the $\star$ symbol.}
    \label{fig-app:results-pareto-aim-acc-xgb-ablation}
\end{figure*}

With this experimental setup, we aim to demonstrate that the main qualitative findings in Sections~\ref{sec:results} and~\ref{sec:discussion} exhibit robustness under this alternative calibration scheme. 
Figure \ref{fig-app:results-pareto-aim-acc-xgb-ablation} presents a comparison of all fairness mechanisms, in which post-processing methods are calibrated using a DP-compliant dataset subject to the same privacy guarantees as the training set ($\varepsilon$-DP). While the influence of DP-calibrated data becomes more pronounced in high-privacy regimes ($\varepsilon<1$), post-processing methods consistently remain closer to the ideal value of zero across both EOD and SPD metrics, while preserving utility performance comparable to \dponly{}. This behaviour corroborates the trends established in Section \ref{sub:results-post}.

\begin{figure*}[!ht]
    \centering
    \centering
    \includegraphics[width=\linewidth]{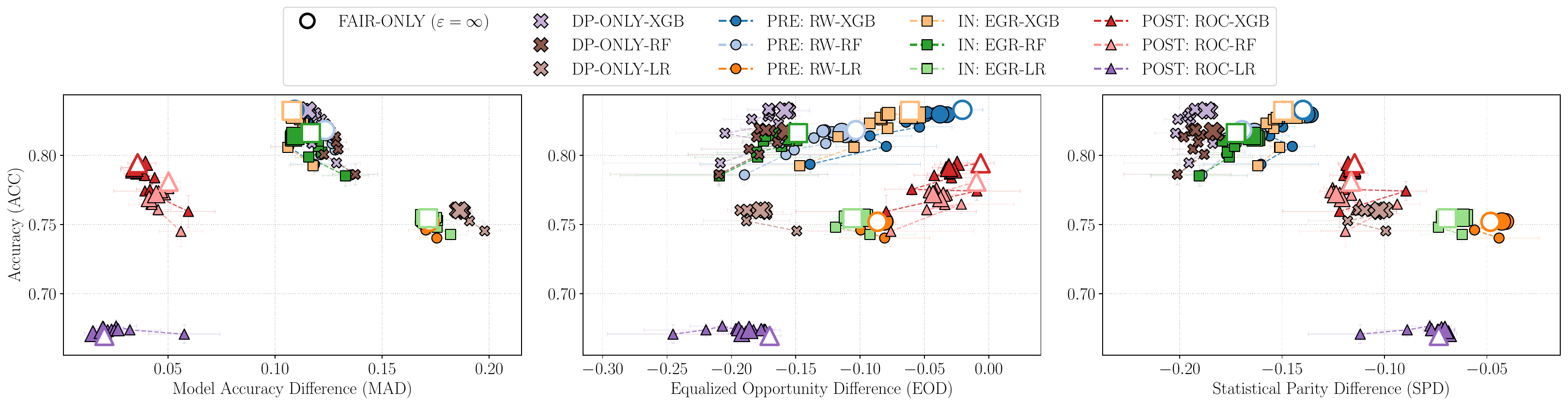}
    \label{fig-app:results-pareto-acc-adult-all-models-ablation}
    \caption{Accuracy--fairness trade-off under the \textbf{AIM} synthesizer across the Adult dataset and three different classifiers (\textbf{Logistic Regression}, \textbf{Random Forest}, and \textbf{XGBoost}). 
    Each subfigure reports accuracy (ACC) versus three fairness metrics (MAD, EOD, SPD) for three different fairness mechanisms, across representative interventions from each stage. 
    The post-processing \dpfair{} settings are calibrated using a DP synthetic calibration set. 
    Marker size encodes the privacy budget $\varepsilon$ for both \dponly{} and \dpfair{} settings, while hollow markers indicate the \faironly{} setting ($\varepsilon=\infty$).
    }
    \label{fig-app:results-pareto-aim-acc-all-models-ablation}
\end{figure*}

Similar to Figure~\ref{fig:results-pareto-aim-acc-all-models} in the main paper, Figure \ref{fig-app:results-pareto-aim-acc-all-models-ablation} reproduces the multi-classifier analysis across the Adult dataset for three learning algorithms, each paired with a representative fairness intervention corresponding to its respective processing stage. 
The geometric configuration and relative ordering of mechanisms in Figure \ref{fig-app:results-pareto-aim-acc-all-models-ablation} remain largely invariant to those of Figure~\ref{fig:results-pareto-aim-acc-all-models}. 
Specifically, these trends persist independently of the calibration set composition, confirming that the adoption of DP-calibrated data does not fundamentally alter the outcome of the fairness correction.

Generally, across the classifiers, ROC and EqOdds continue to provide the strongest overall fairness--utility trade-offs among the evaluated interventions, even when fairness post-processing is calibrated without direct access to non-private calibration records. 
This indicates that the advantage of post-processing observed in the main experiments is not solely driven by the use of a real held-out calibration set.

\subsection{Ablation: Hyperparameter Sensitivity for In-Processing Methods} \label{app:ablation_in_processing}

In Section~\ref{sub:overview_benchmark}, the benchmark evaluates each fairness mechanism using its default hyperparameter configuration. As discussed previously in Section~\ref{sub:results-in}, several methods yielded only modest and inconsistent improvements in fairness while largely preserving predictive utility. This trend was particularly pronounced for the in-processing mechanisms. To determine the extent to which these results may be influenced by the use of default hyperparameters, we perform an ablation study in which multiple hyperparameter combinations are evaluated for both the EGR and GSR mechanisms. Finally, we compare the configurations that achieve the most favourable utility--fairness trade-off for each fairness metric.

Hyperparameters were selected based on their expected influence on model performance, while keeping the fairness constraints and fairness-violation bounds fixed. For EGR, we vary \textit{max\_iter} and \textit{nu}. The parameter \textit{max\_iter} determines the maximum number of optimisation iterations, whereas \textit{nu} specifies the convergence threshold. For GSR, we vary \textit{grid\_size} and \textit{grid\_limit}. The former controls the size of the search space, while the latter determines the range of Lagrange multipliers considered during optimisation.

For each mechanism, we construct a grid of 25 hyperparameter combinations. The full set of candidate values is reported in Table~\ref{tab-app:fairness_hyperparameter_grid}. For each fairness mechanism, we then identify the configuration that provides the best utility--fairness trade-off as per Equation~\eqref{eq:objective_weighted_euclidean_dist} and visualise the corresponding results.

\begin{table}[htbp] 
\centering 
\begin{tabular}{lll} 
\hline \textbf{Algorithm} & \textbf{Hyperparameter} & \textbf{Candidate Values} \\
\hline 
EGR & \texttt{max\_iter} & \{50, 75, 100, 125, 150\} \\ 
EGR & \texttt{nu} & \{None, 0.01, 0.05, 0.1, 0.5\}   \\ 
GSR & \texttt{grid\_size} & \{10, 15, 20, 30, 50\}   \\ 
GSR & \texttt{grid\_limit} & \{2.0, 1.0, 1.5, 2.5, 3.0\} \\ 
\hline 
\end{tabular} 
\caption{Hyperparameter grid used for tuning fairness-aware reduction methods.} 
\label{tab-app:fairness_hyperparameter_grid} 
\end{table}

\begin{figure*}[!ht]
    \centering
    \centering
    \includegraphics[width=\linewidth]{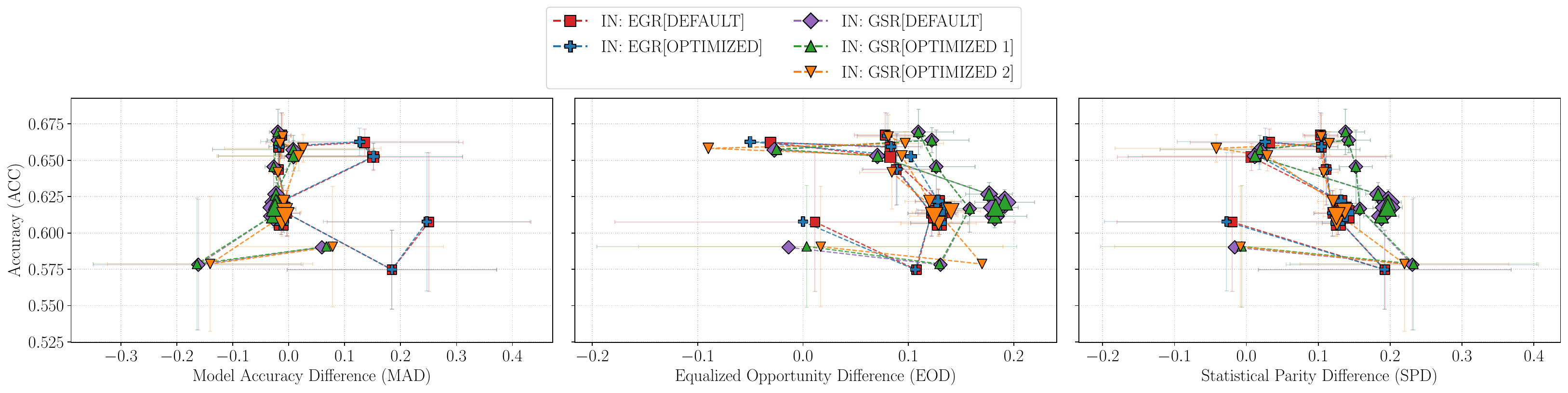}
    \label{fig-app:in-proc-ablation-compas}
    \caption{Accuracy--fairness trade-off under the \textbf{AIM} synthesiser across the Compas dataset, two fairness mechanisms and the classifier \textbf{XGBoost}. 
    Each subfigure reports accuracy (ACC) versus three fairness metrics (MAD, EOD, SPD) for 2 different fairness mechanisms, across 5 hyperparameter combinations. 
    Marker size encodes the privacy budget $\varepsilon$ for both \dponly{} and \dpfair{} settings, while hollow markers indicate the \faironly{} setting ($\varepsilon=\infty$).
    }
    \label{fig-app:in-proc-ablation}
\end{figure*}

The results presented in Figure~\ref{fig-app:in-proc-ablation} indicate that variations in hyperparameter configuration lead to slight changes in the utility--fairness trade-off. However, these differences remain modest and consistent across configurations. This suggests that the trends identified in Section~\ref{sub:results-in} are robust and remain largely unchanged even after hyperparameter tuning of the fairness mechanisms.

\subsection{Additional Results} \label{app:add_results}

In this section, we report additional experimental results that complement those presented in Section~\ref{sec:results}. Specifically, we consider the following combinations of classifiers and DP synthesizers:
\begin{enumerate*}
    \item XGBoost with MST;
    \item Logistic Regression with AIM;
    \item Random Forest with AIM,
\end{enumerate*}
\noindent which are reported in Subsections~\ref{app:results_xgb},~\ref{app:results_lr}, and~\ref{app:results_rf}, respectively.
In addition, Subsection~\ref{app:results_bod} presents a comparative analysis of the different classifier configurations on the BiasOnDemand dataset. Along with this analysis, we present a summary of the common findings and whether the patterns found are either classifier-dependent or dataset-dependent in Subsection \ref{app:correlation}.

\subsubsection{Results of XGBoost with MST}
\label{app:results_xgb}

\begin{figure*}[!ht]
    \centering
    \begin{subfigure}{0.99\textwidth}
        \centering
        \includegraphics[width=\linewidth]{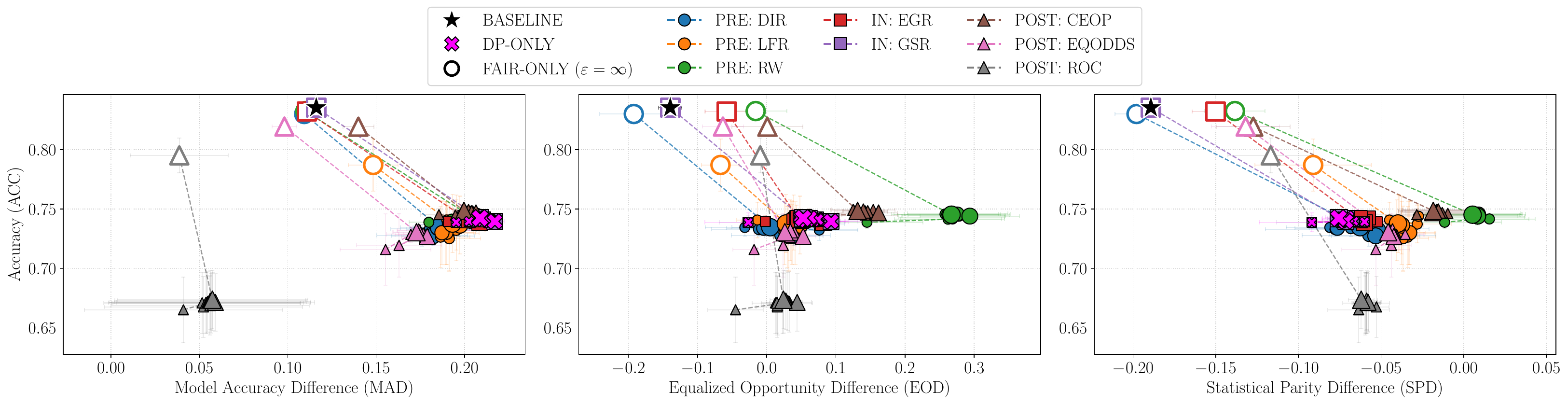}
        \caption{\textbf{Adult}.}
        \label{fig-app:results-pareto-acc-adult-xgb}
    \end{subfigure}\\
    \hfill
    \begin{subfigure}{0.99\textwidth}
        \centering
        \includegraphics[width=\linewidth]{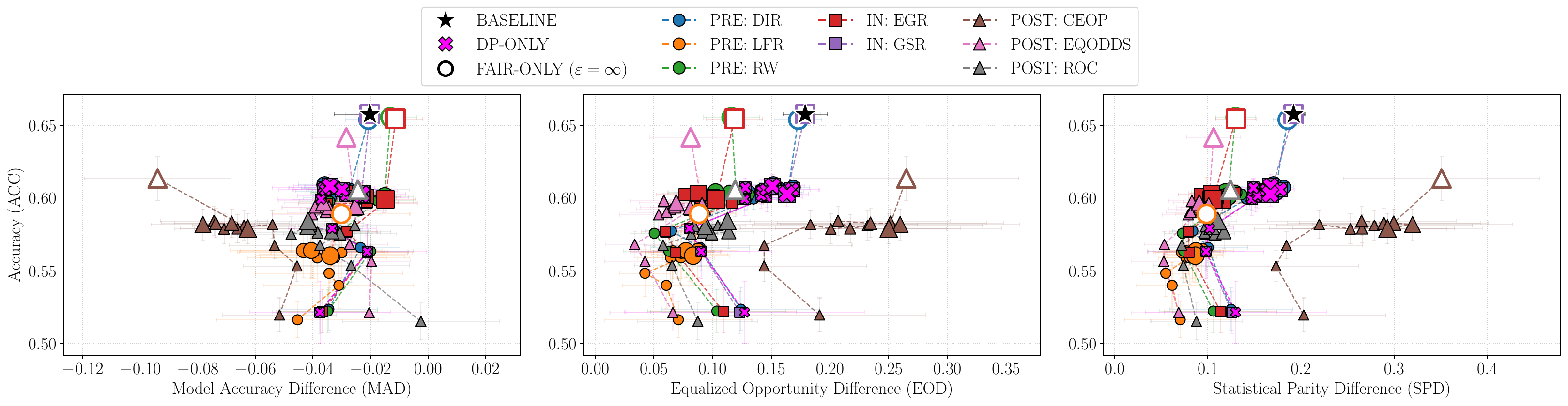}
        \caption{\textbf{COMPAS}.}
        \label{fig-app:results-pareto-acc-compas-xgb}
    \end{subfigure}\\
    \begin{subfigure}{0.99\textwidth}
        \centering
        \includegraphics[width=\linewidth]{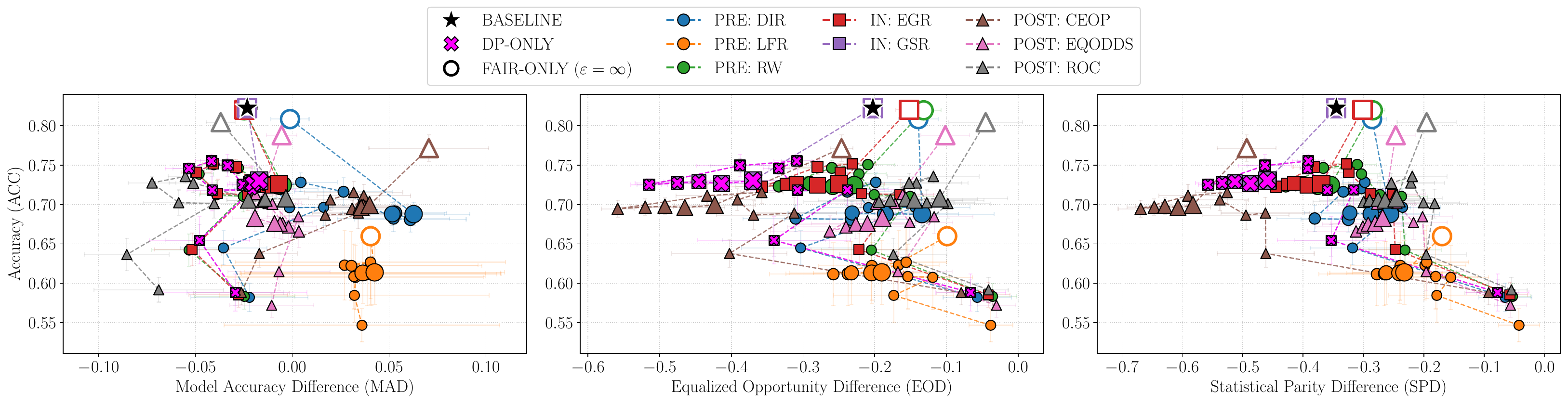}
        \caption{\textbf{ACSIncome}.}
        \label{fig-app:results-pareto-acc-acsincome-xgb}
    \end{subfigure}\\
    \hfill
    \begin{subfigure}{0.99\textwidth}
        \centering
        \includegraphics[width=\linewidth]{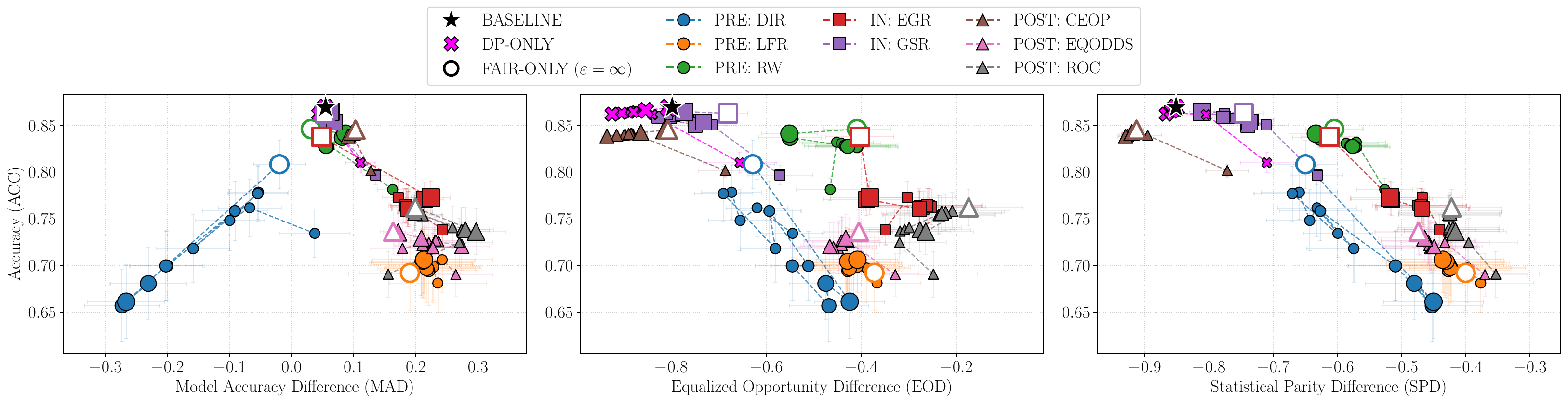}
        \caption{\textbf{BiasOnDemand (Configuration 5)}.}
        \label{fig-app:results-pareto-acc-bod-xgb}
    \end{subfigure}
    \caption{Accuracy--fairness trade-off under the \textbf{MST} synthesizer across four datasets with the \textbf{XGBoost} classifier. Each subfigure reports accuracy (ACC) versus three fairness metrics (MAD, EOD, SPD) for each fairness intervention applied at the pre-, in-, and post-processing stages. 
    Marker size encodes the privacy budget $\varepsilon$ for both \dponly{} and \dpfair{} settings, while hollow markers indicate the \faironly{} setting ($\varepsilon=\infty$).
    The \baseline{} setting (no DP and no fairness intervention) is indicated by the $\star$ symbol.}
    \label{fig-app:results-pareto-mst-acc-xgb}
\end{figure*}

Figure~\ref{fig-app:results-pareto-mst-acc-xgb} reports the results obtained with the XGBoost classifier trained on data generated by the MST synthesizer.
Overall, post-processing methods, in particular ROC and EqOdds, achieve the most consistent reductions in group disparities across both EOD and SPD, while maintaining utility relatively close to the \dponly{} baseline.
Other mechanisms, such as LFR (pre-processing) and EGR (in-processing), can also reduce disparities, but do so less favorably: LFR consistently incurs substantial utility loss, whereas EGR exhibits more limited and privacy-budget-dependent improvements when compared to post-processing.
Among pre-processing methods, Reweighing emerges as the most reliable option in this setting. RW provides noticeable improvements in SPD while keeping accuracy close to the \dponly{} configuration. However, these gains may come at the expense of other fairness notions, such as EOD, as observed for the Adult dataset.
Taken together, the XGBoost $\times$ MST results highlight two main observations: (i) when predictive utility is a priority, aggressive pre-processing methods such as LFR are not suitable; and (ii) in-processing approaches offer more limited benefits compared to pre- and post-processing, with post-processing methods providing the most favorable and robust fairness--utility trade-offs under this setting.

\subsubsection{Results of Logistic Regression with AIM}
\label{app:results_lr}
\begin{figure*}[!ht]
    \centering
    \begin{subfigure}{0.99\textwidth}
        \centering
        \includegraphics[width=\linewidth]{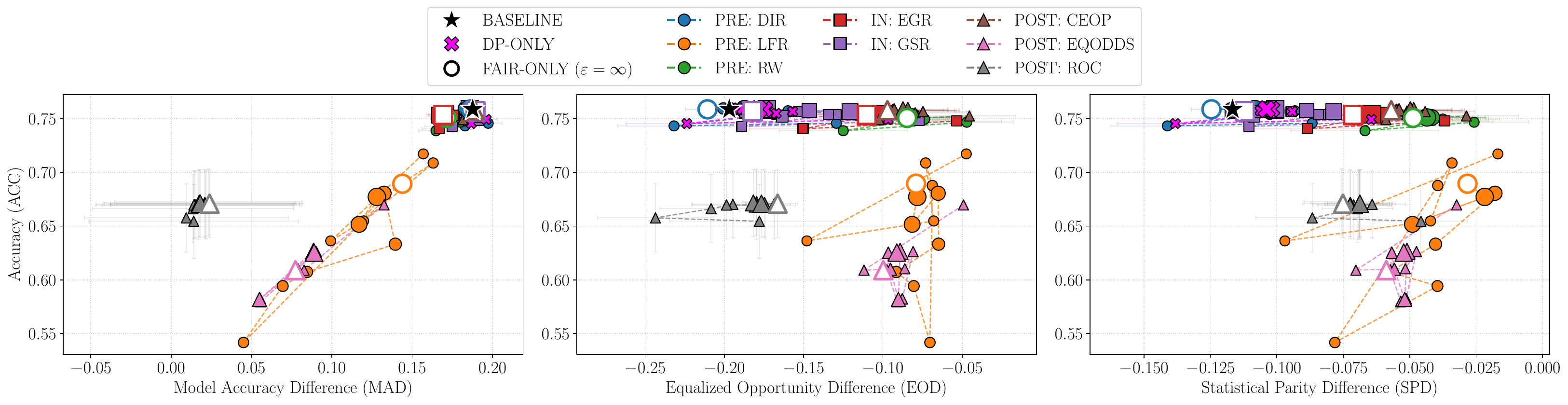}
        \caption{\textbf{Adult}.}
        \label{fig-app:results-pareto-acc-adult-lr}
    \end{subfigure}\\
    \hfill
    \begin{subfigure}{0.99\textwidth}
        \centering
        \includegraphics[width=\linewidth]{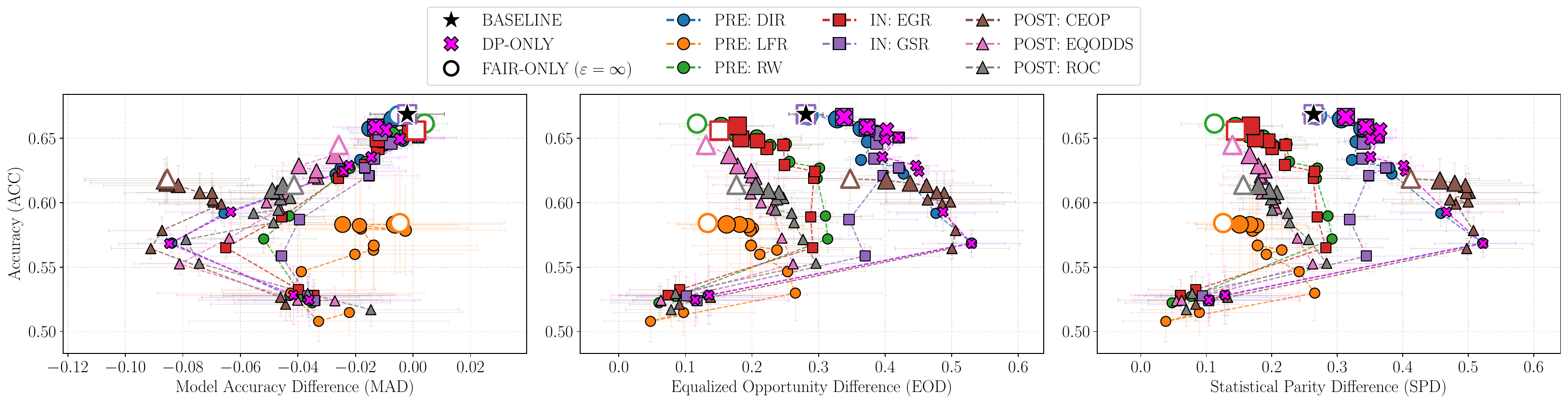}
        \caption{\textbf{COMPAS}.}
        \label{fig-app:results-pareto-acc-compas-lr}
    \end{subfigure}\\
    \begin{subfigure}{0.99\textwidth}
        \centering
        \includegraphics[width=\linewidth]{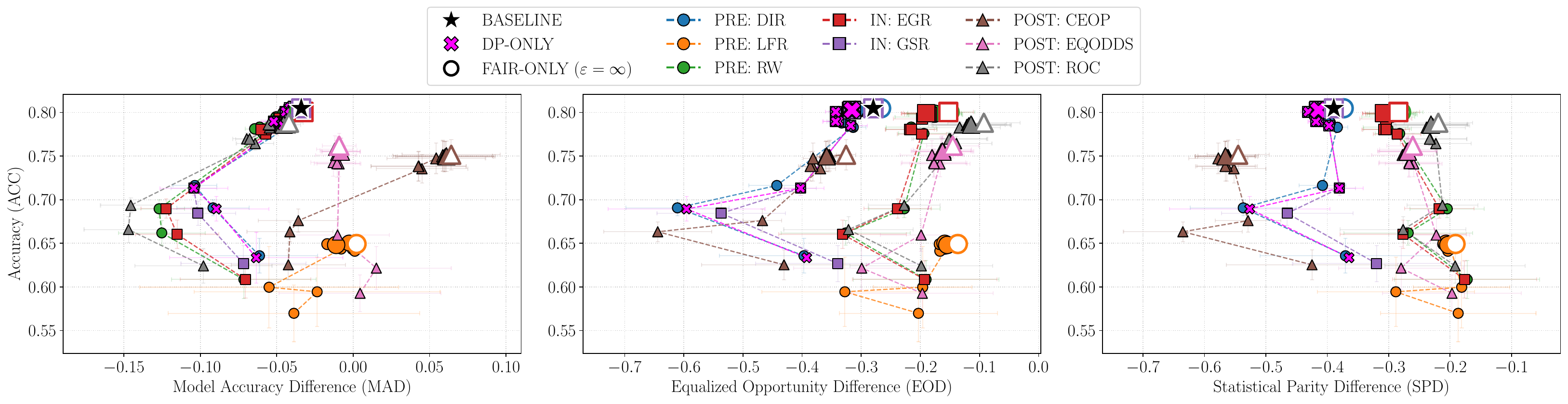}
        \caption{\textbf{ACSIncome}.}
        \label{fig-app:results-pareto-acc-acsincome-lr}
    \end{subfigure}\\
    \hfill
    \begin{subfigure}{0.99\textwidth}
        \centering
        \includegraphics[width=\linewidth]{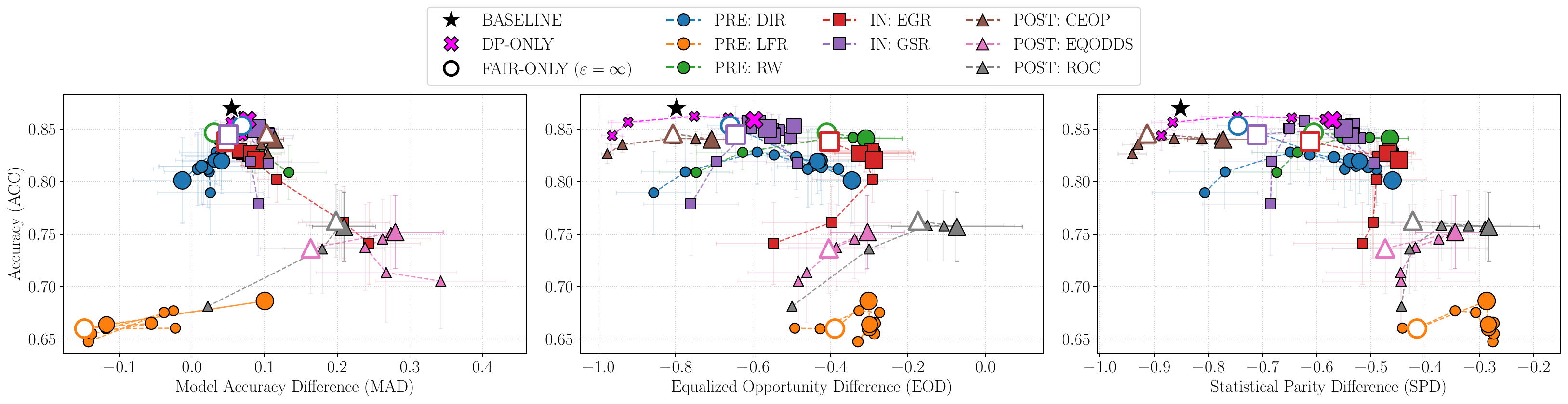}
        \caption{\textbf{BiasOnDemand (Configuration 5)}.}
        \label{fig-app:results-pareto-acc-bod-lr}
    \end{subfigure}
    \caption{Accuracy--fairness trade-off under the \textbf{AIM} synthesizer across four datasets with the \textbf{Logistic Regression} classifier. Each subfigure reports accuracy (ACC) versus three fairness metrics (MAD, EOD, SPD) for each fairness intervention applied at the pre-, in-, and post-processing stages. 
    Marker size encodes the privacy budget $\varepsilon$ for both \dponly{} and \dpfair{} settings, while hollow markers indicate the \faironly{} setting ($\varepsilon=\infty$).
    The \baseline{} setting (no DP and no fairness intervention) is indicated by the $\star$ symbol.}
    \label{fig-app:results-pareto-aim-acc-lr}
\end{figure*}

\begin{figure*}[!ht]
    \centering
    \begin{subfigure}{0.99\textwidth}
        \centering
        \includegraphics[width=\linewidth]{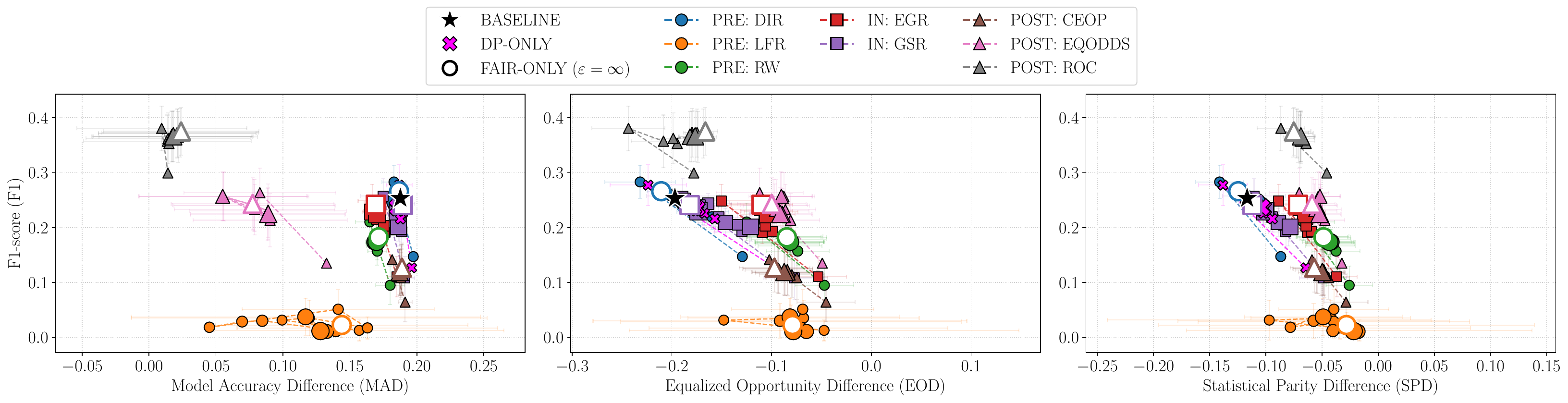}
        \caption{\textbf{Adult}.}
        \label{fig-app:results-pareto-f1-adult-lr}
    \end{subfigure}\\
    \hfill
    \begin{subfigure}{0.99\textwidth}
        \centering
        \includegraphics[width=\linewidth]{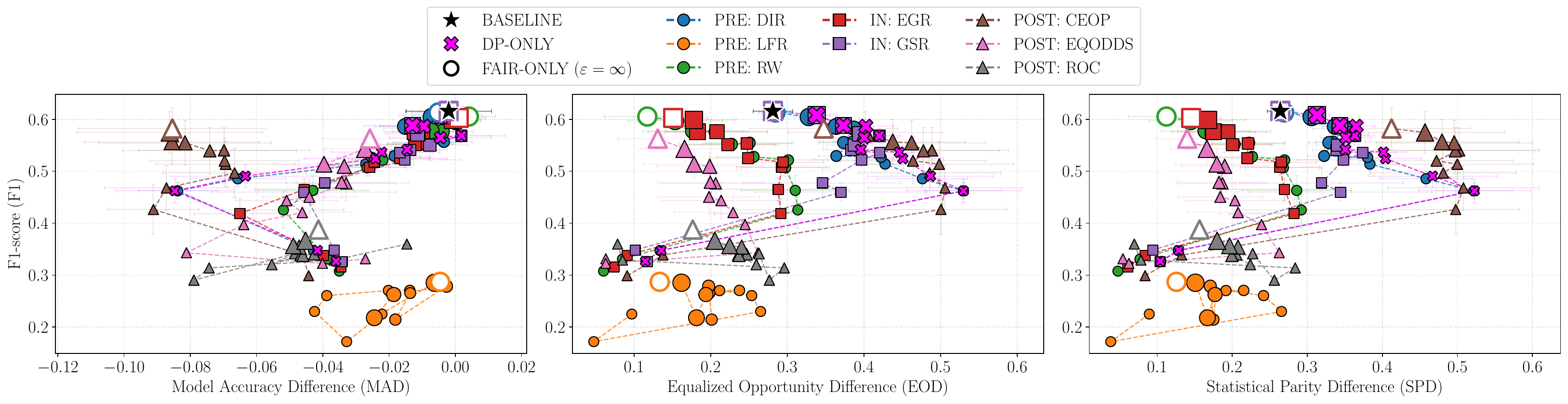}
        \caption{\textbf{COMPAS}.}
        \label{fig-app:results-pareto-f1-compas-lr}
    \end{subfigure}\\
    \begin{subfigure}{0.99\textwidth}
        \centering
        \includegraphics[width=\linewidth]{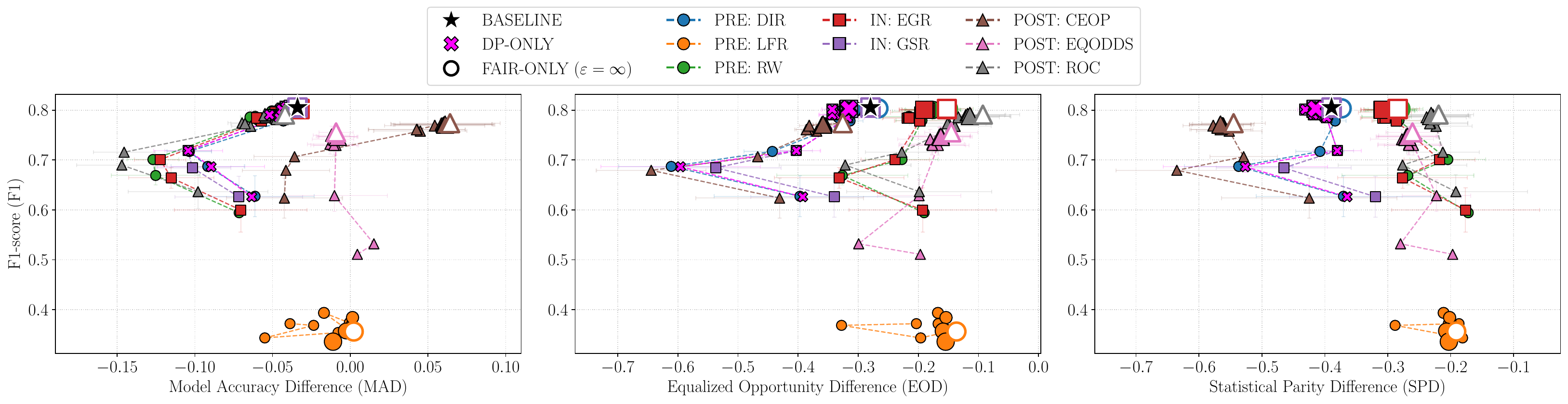}
        \caption{\textbf{ACSIncome}.}
        \label{fig-app:results-pareto-f1-acsincome-lr}
    \end{subfigure}\\
    \hfill
    \begin{subfigure}{0.99\textwidth}
        \centering
        \includegraphics[width=\linewidth]{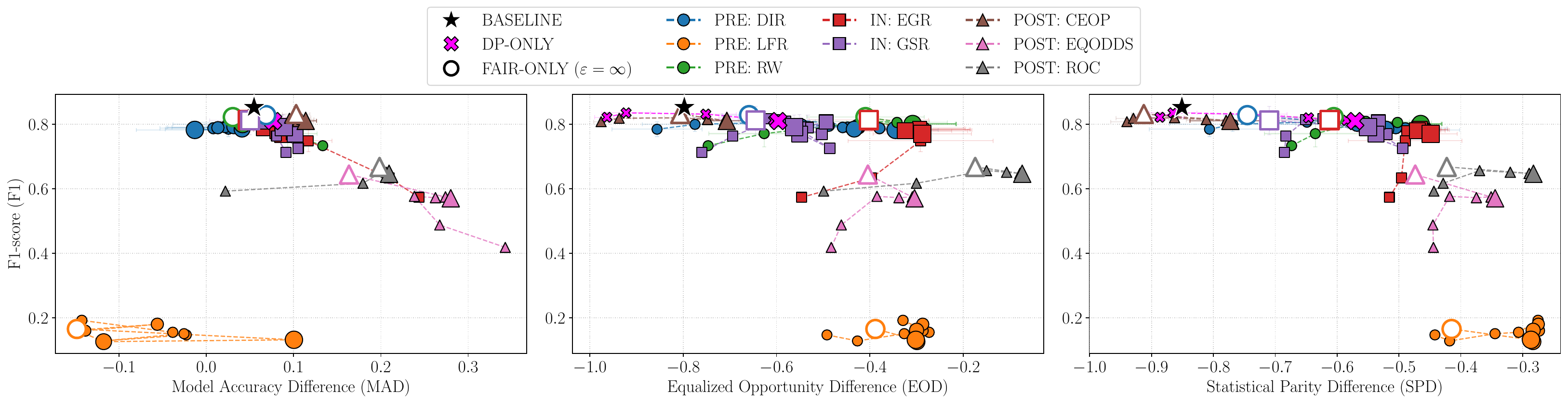}
        \caption{\textbf{BiasOnDemand (Configuration 5)}.}
        \label{fig-app:results-pareto-f1-bod-lr}
    \end{subfigure}
    \caption{F1--fairness trade-off under the \textbf{AIM} synthesizer across four datasets with the \textbf{Logistic Regression} classifier. The figure reports F1 versus three fairness metrics (MAD, EOD, SPD) for each fairness intervention applied at the pre-, in-, and post-processing stages. 
    Marker size encodes the privacy budget $\varepsilon$ for both \dponly{} and \dpfair{} settings, while hollow markers indicate the \faironly{} setting ($\varepsilon=\infty$).
    The \baseline{} setting (no DP and no fairness intervention) is indicated by the $\star$ symbol.}
    \label{fig-app:f1-adult-lr-aim}
\end{figure*}

Figure~\ref{fig-app:results-pareto-aim-acc-lr} reports the results obtained with Logistic Regression (LR) trained on AIM synthetic data.
Compared to the XGBoost-based results in Section~\ref{sec:results}, some fairness mechanisms exhibit a less favorable accuracy--fairness trade-off on the Adult dataset.
At first glance, this suggests a stronger dependence on the choice of classifier.
A closer inspection, however, reveals that this behavior is largely driven by differences in \emph{error profiles} rather than a failure of the fairness mechanisms.
In particular, the baseline LR model achieves relatively high accuracy but suffers from low recall on the Adult dataset, indicating conservative predictions.
Several fairness interventions reduce accuracy while improving recall, leading to a more balanced precision--recall trade-off.
This effect is captured by the F1-score results in Figure~\ref{fig-app:f1-adult-lr-aim}, where the apparent degradation observed under accuracy largely disappears.
These observations highlight that the perceived effectiveness of fairness interventions can depend on the chosen utility metric and on the calibration properties of the underlying classifier.
Importantly, despite these metric-dependent effects, the overall qualitative patterns remain consistent with the main results:
Reweighing, Exponentiated Gradient Reduction, Reject Option Classification, and Equalized Odds Post-Processing continue to reduce group disparities while maintaining utility close to the \dponly{} baseline.
This confirms that the main conclusions that our benchmark extends beyond XGBoost and remain valid for linear classifiers such as Logistic Regression.

\subsubsection{Results of Random Forest with AIM}
\label{app:results_rf}

\begin{figure*}[!ht]
    \centering
    \begin{subfigure}{0.99\textwidth}
        \centering
        \includegraphics[width=\linewidth]{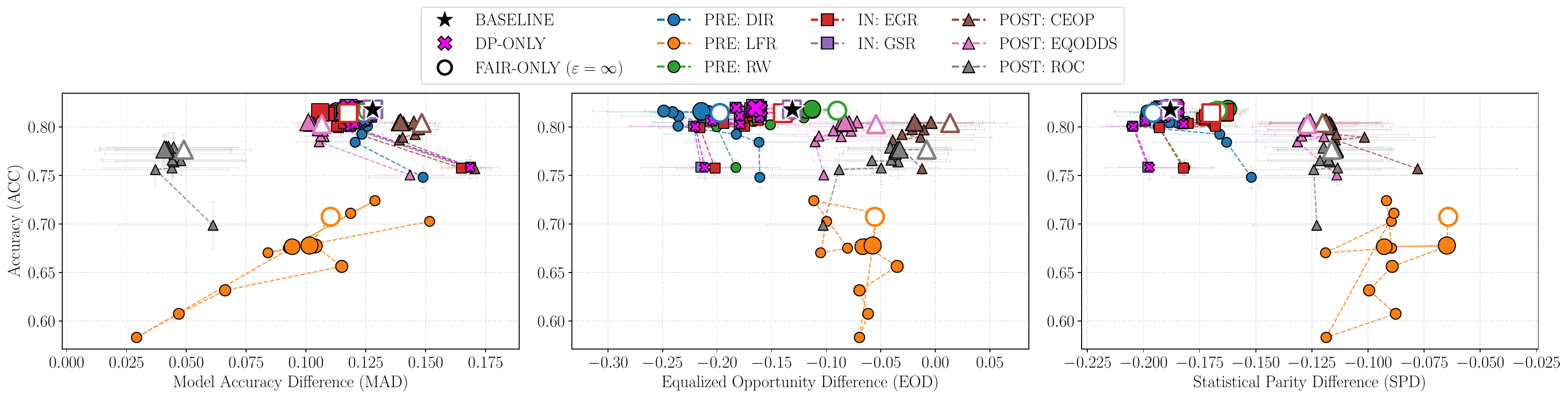}
        \caption{\textbf{Adult}.}
        \label{fig-app:results-pareto-acc-adult-rf}
    \end{subfigure}\\
    \hfill
    \begin{subfigure}{0.99\textwidth}
        \centering
        \includegraphics[width=\linewidth]{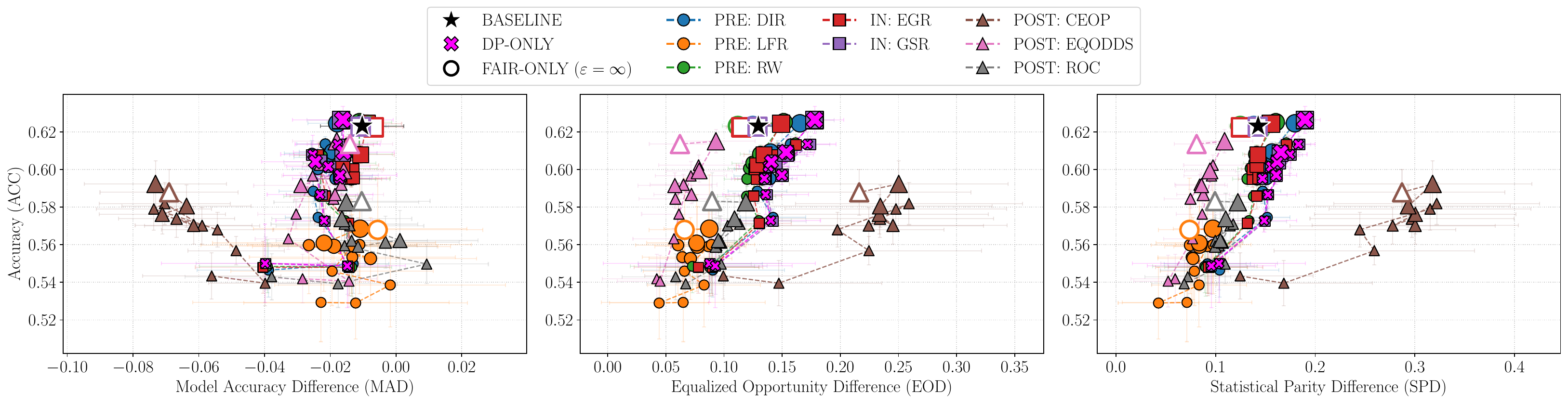}
        \caption{\textbf{COMPAS}.}
        \label{fig-app:results-pareto-acc-compas-rf}
    \end{subfigure}\\
    \begin{subfigure}{0.99\textwidth}
        \centering
        \includegraphics[width=\linewidth]{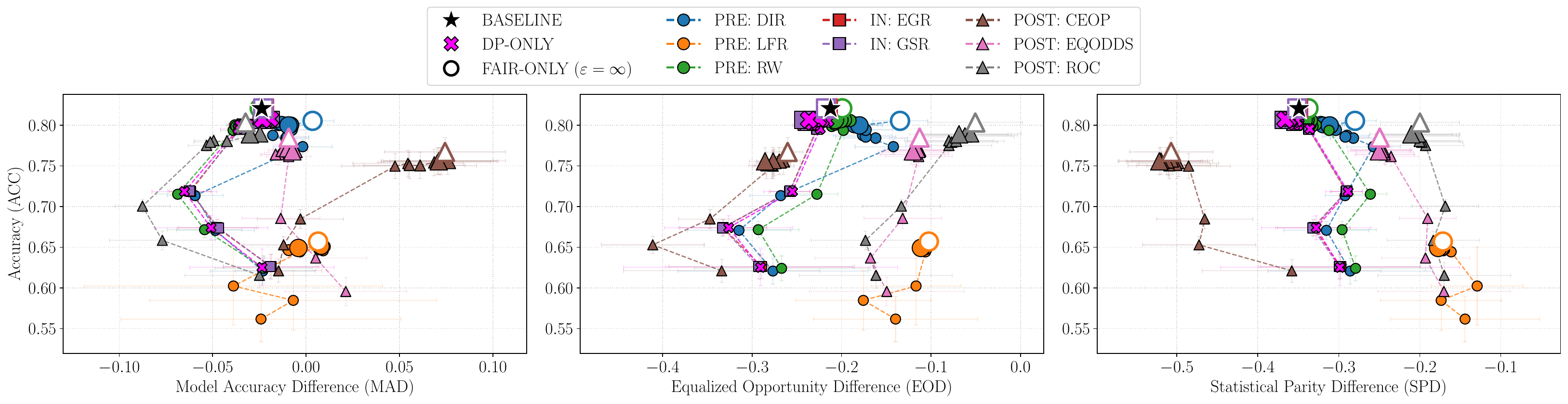}
        \caption{\textbf{ACSIncome}.}
        \label{fig-app:results-pareto-acc-acsincome-rf}
    \end{subfigure}\\
    \hfill
    \begin{subfigure}{0.99\textwidth}
        \centering
        \includegraphics[width=\linewidth]{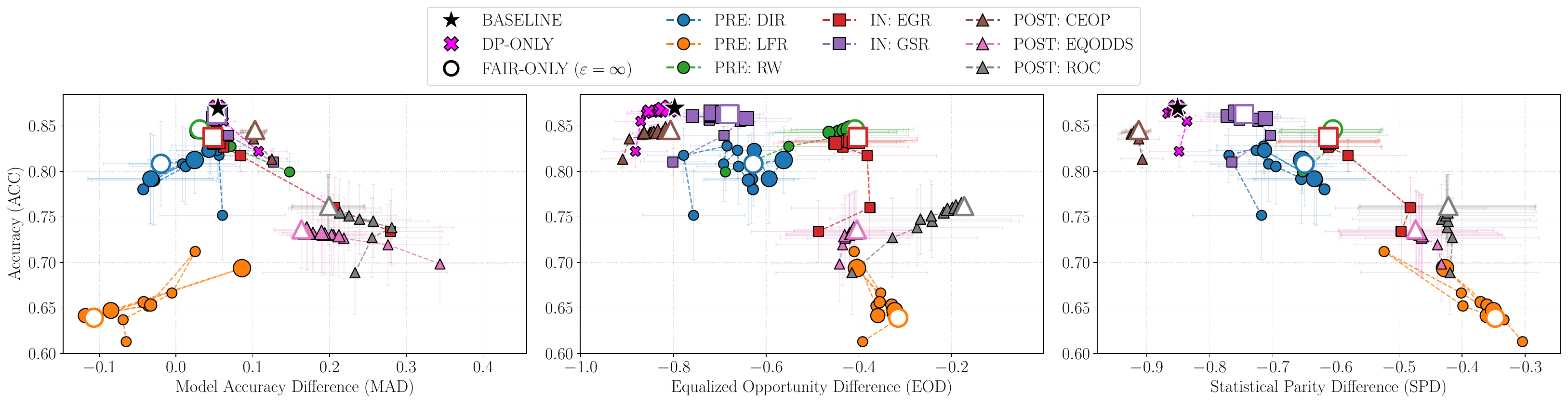}
        \caption{\textbf{BiasOnDemand (Configuration 5)}.}
        \label{fig-app:results-pareto-acc-bod-rf}
    \end{subfigure}
    \caption{Accuracy--fairness trade-off under the \textbf{AIM} synthesizer across four datasets with the \textbf{Random Forest} classifier. Each subfigure reports accuracy (ACC) versus three fairness metrics (MAD, EOD, SPD) for each fairness intervention applied at the pre-, in-, and post-processing stages. 
    Marker size encodes the privacy budget $\varepsilon$ for both \dponly{} and \dpfair{} settings, while hollow markers indicate the \faironly{} setting ($\varepsilon=\infty$).
    The \baseline{} setting (no DP and no fairness intervention) is indicated by the $\star$ symbol.}
    \label{fig-app:results-pareto-aim-acc-rf}
\end{figure*}

Figure~\ref{fig-app:results-pareto-aim-acc-rf} reports the results obtained using the Random Forest (RF) classifier with the AIM synthesizer. Overall, the trends identified in Section~\ref{sec:results} remain consistent under this configuration. In particular, post-processing methods (ROC and EqOdds) continue to offer the most reliable fairness--utility trade-offs: across datasets and privacy budgets, \dpfair{} models with post-processing closely track the utility of \baseline{} and \dponly{}, while substantially reducing group disparities. The isolated analyses further show that, although models trained on DP synthetic data may initially exhibit worse fairness metrics than the \baseline{}, several mechanisms consistently recover fairness without significant additional utility loss. Specifically, ROC and EqOdds provide the strongest and most stable corrections, while Reweighing and Exponentiated Gradient Reduction also achieve meaningful disparity reductions with moderate utility impact. By contrast, improvements observed for other mechanisms (\eg, GSR) tend to be dataset-specific and do not generalize across domains. Taken together, these results confirm that the qualitative conclusions drawn in the main paper are robust to the choice of classifier: under AIM, post-processing methods—particularly ROC and EqOdds—consistently deliver the best fairness--utility trade-offs, a pattern that holds across datasets and learning algorithms.

\subsubsection{Results of Bias on Demand}
\label{app:results_bod}

\begin{figure*}[!ht]
    \centering
    \begin{subfigure}{0.99\textwidth}
        \centering
        \includegraphics[width=\linewidth]{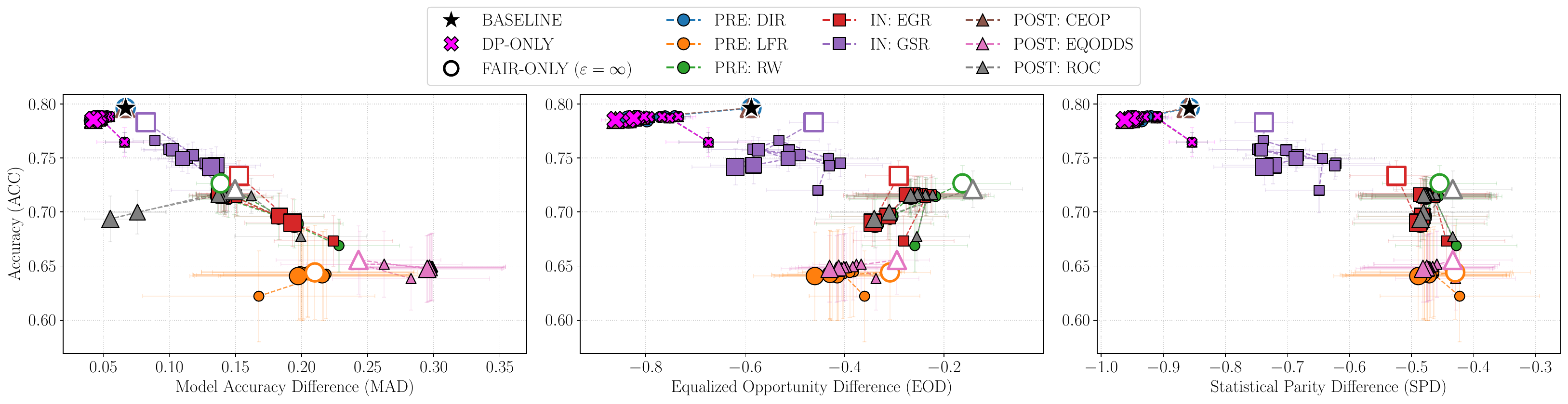}
        \caption{\textbf{BiasOnDemand Configuration 1}.}
        \label{fig-app:results-pareto-acc-bod-1-xgb}
    \end{subfigure}\\
    \hfill
    \begin{subfigure}{0.99\textwidth}
        \centering
        \includegraphics[width=\linewidth]{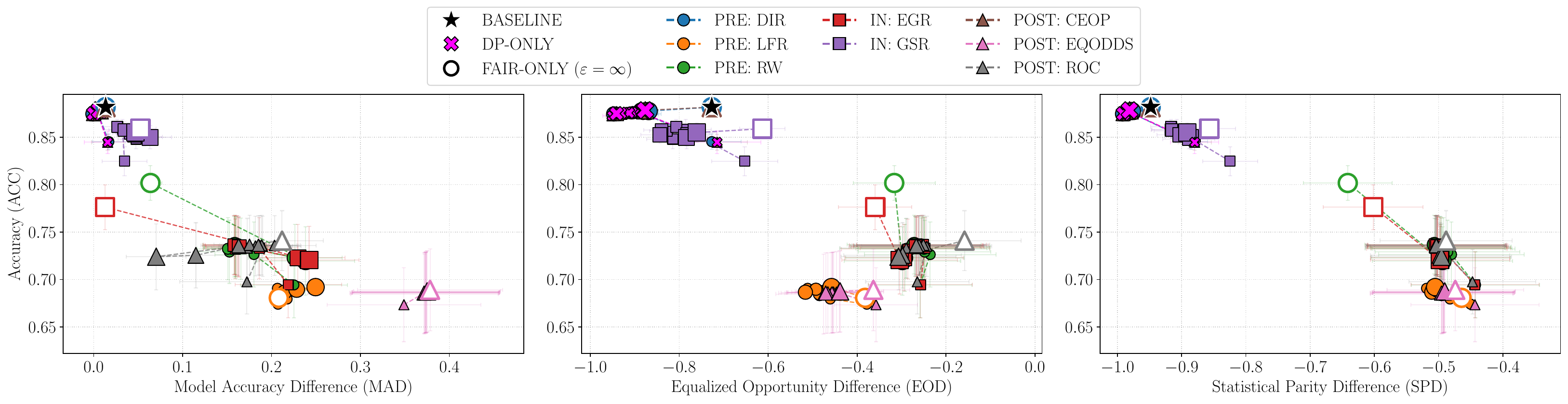}
        \caption{\textbf{BiasOnDemand Configuration 2}.}
        \label{fig-app:results-pareto-acc-bod-2-xgb}
    \end{subfigure}\\
    \begin{subfigure}{0.99\textwidth}
        \centering
        \includegraphics[width=\linewidth]{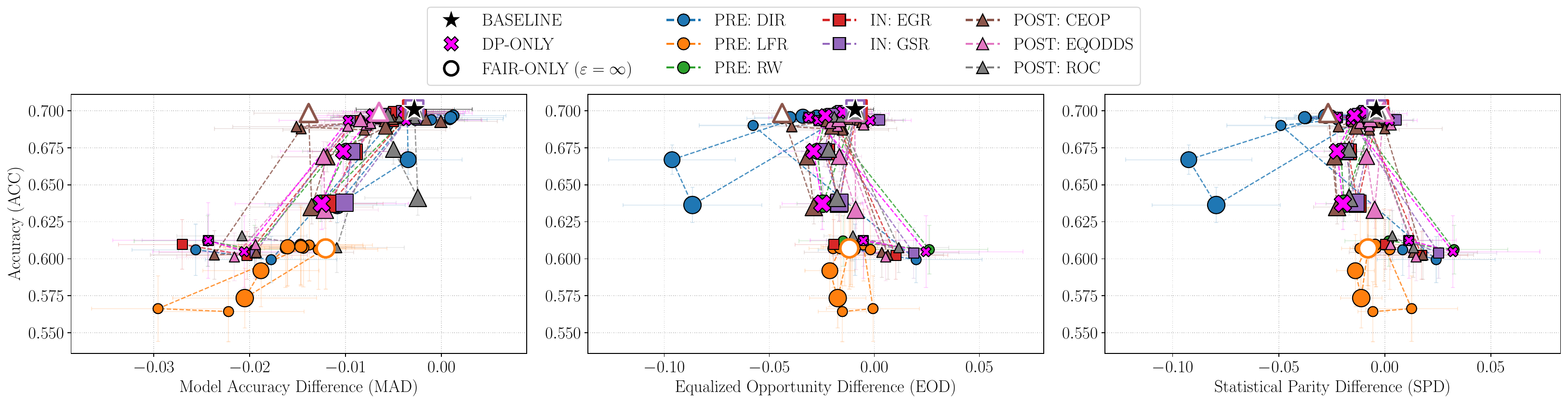}
        \caption{\textbf{BiasOnDemand Configuration 3}.}
        \label{fig-app:results-pareto-acc-bod-3-xgb}
    \end{subfigure}\\
    \hfill
    \caption{Accuracy--fairness trade-off under the \textbf{MST} synthesizer across configurations 1, 2, and 3 of the BoD dataset with the \textbf{XGBoost} classifier. Each subfigure reports accuracy (ACC) versus three fairness metrics (MAD, EOD, SPD) for each fairness intervention applied at the pre-, in-, and post-processing stages. 
    Marker size encodes the privacy budget $\varepsilon$ for both \dponly{} and \dpfair{} settings, while hollow markers indicate the \faironly{} setting ($\varepsilon=\infty$).
    The \baseline{} setting (no DP and no fairness intervention) is indicated by the $\star$ symbol.}
    \label{fig-app:results-pareto-mst-acc-bod-123-xgb}
\end{figure*}

\begin{figure*}[!ht]
    \centering
    \begin{subfigure}{0.99\textwidth}
        \centering
        \includegraphics[width=\linewidth]{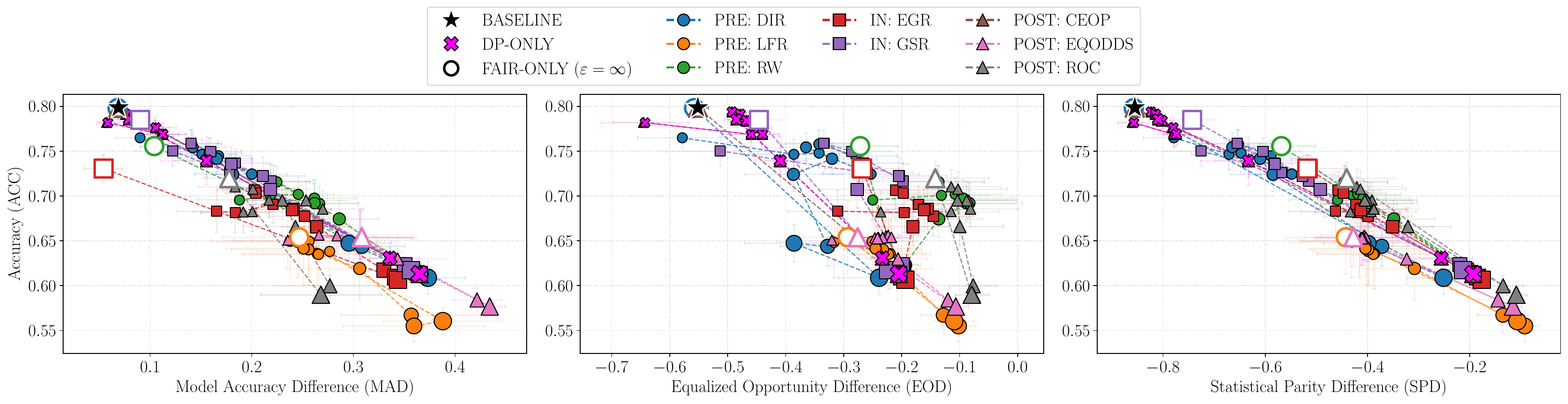}
        \caption{\textbf{BiasOnDemand Configuration 4}.}
        \label{fig-app:results-pareto-acc-bod-4-xgb}
    \end{subfigure}\\
    \hfill
    \begin{subfigure}{0.99\textwidth}
        \centering
        \includegraphics[width=\linewidth]{figures/appendix/fig_results_XGB_mst_ACC_all_BoD-5.pdf}
        \caption{\textbf{BiasOnDemand (Configuration 5)}.}
        \label{fig-app:results-pareto-acc-bod-5-xgb}
    \end{subfigure}\\
    \begin{subfigure}{0.99\textwidth}
        \centering
        \includegraphics[width=\linewidth]{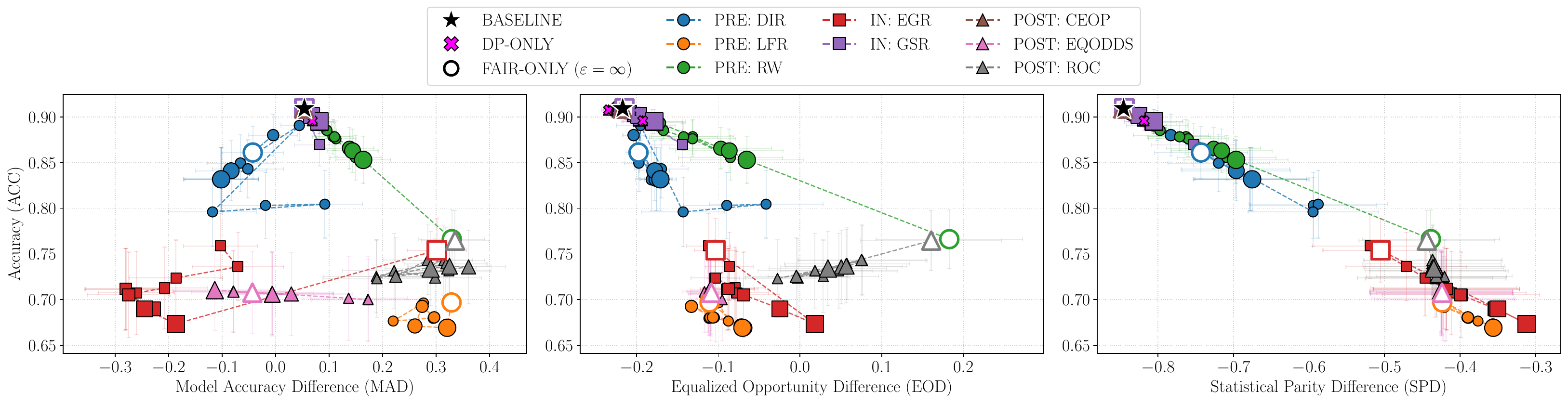}
        \caption{\textbf{BiasOnDemand Configuration 6}.}
        \label{fig-app:results-pareto-acc-bod-6-xgb}
    \end{subfigure}\\
    \hfill
    \caption{Accuracy--fairness trade-off under the \textbf{MST} synthesizer across configurations 4, 5, and 6 of the BoD dataset with the \textbf{XGBoost} classifier. Each subfigure reports accuracy (ACC) versus three fairness metrics (MAD, EOD, SPD) for each fairness intervention applied at the pre-, in-, and post-processing stages. 
    Marker size encodes the privacy budget $\varepsilon$ for both \dponly{} and \dpfair{} settings, while hollow markers indicate the \faironly{} setting ($\varepsilon=\infty$).
    The \baseline{} setting (no DP and no fairness intervention) is indicated by the $\star$ symbol.}
    \label{fig-app:results-pareto-mst-acc-bod-456-xgb}
\end{figure*}

\begin{figure*}[!ht]
    \centering
    \begin{subfigure}{0.99\textwidth}
        \centering
        \includegraphics[width=\linewidth]{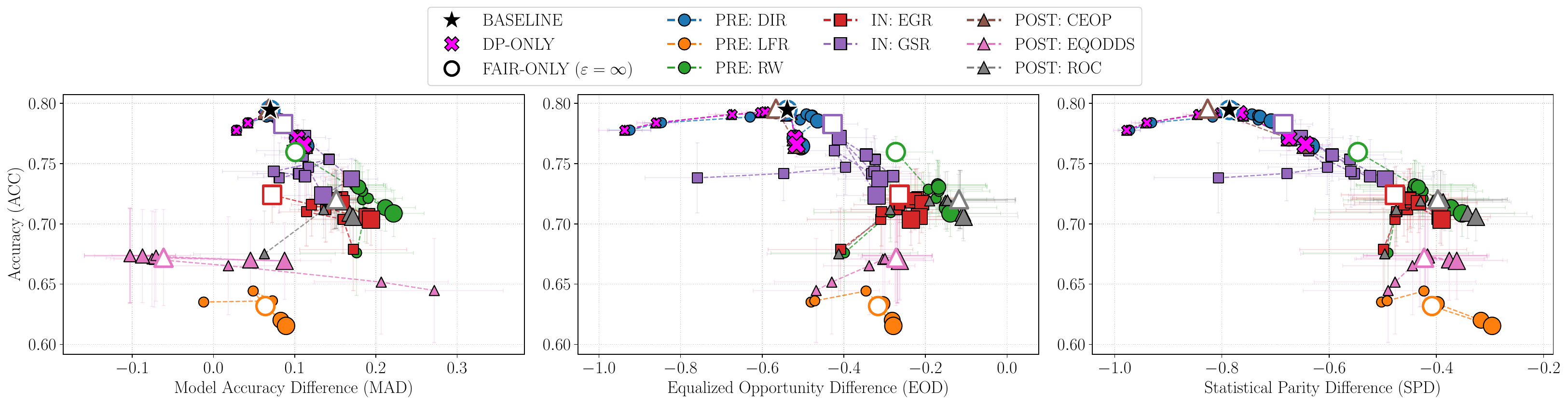}
        \caption{\textbf{BiasOnDemand Configuration 1}.}
        \label{fig-app:results-pareto-acc-bod-1-lr}
    \end{subfigure}\\
    \hfill
    \begin{subfigure}{0.99\textwidth}
        \centering
        \includegraphics[width=\linewidth]{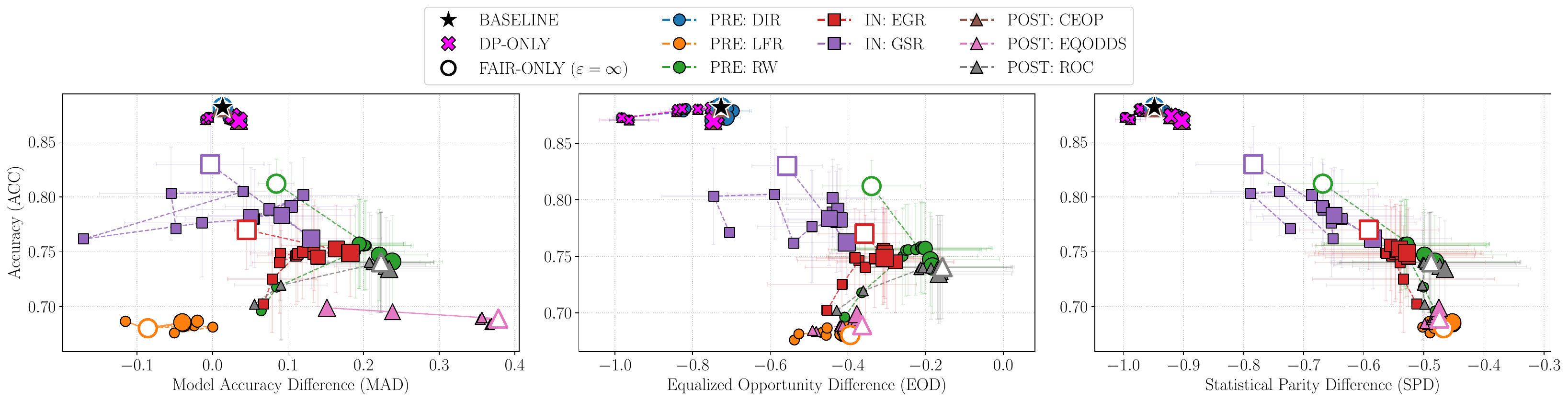}
        \caption{\textbf{BiasOnDemand Configuration 2}.}
        \label{fig-app:results-pareto-acc-bod-2-lr}
    \end{subfigure}\\
    \begin{subfigure}{0.99\textwidth}
        \centering
        \includegraphics[width=\linewidth]{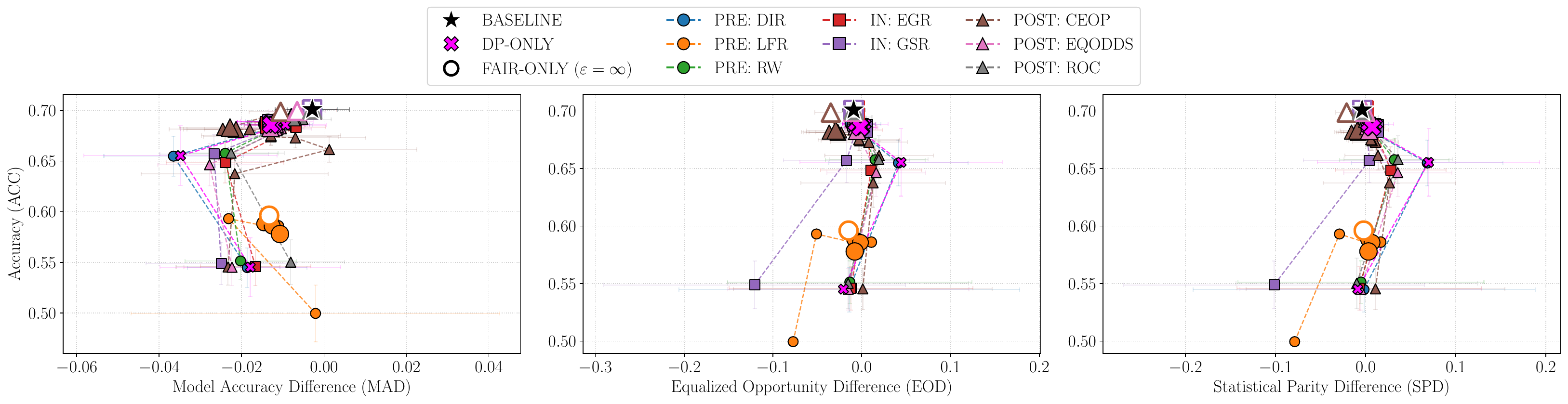}
        \caption{\textbf{BiasOnDemand Configuration 3}.}
        \label{fig-app:results-pareto-acc-bod-3-lr}
    \end{subfigure}\\
    \hfill
    \caption{Accuracy--fairness trade-off under the \textbf{AIM} synthesizer across configurations 1, 2, and 3 of the BoD dataset with the \textbf{Logistic Regression} classifier. Each subfigure reports accuracy (ACC) versus three fairness metrics (MAD, EOD, SPD) for each fairness intervention applied at the pre-, in-, and post-processing stages. 
    Marker size encodes the privacy budget $\varepsilon$ for both \dponly{} and \dpfair{} settings, while hollow markers indicate the \faironly{} setting ($\varepsilon=\infty$).
    The \baseline{} setting (no DP and no fairness intervention) is indicated by the $\star$ symbol.}
    \label{fig-app:results-pareto-aim-acc-bod-123-lr}
\end{figure*}

\begin{figure*}[!ht]
    \centering
    \begin{subfigure}{0.99\textwidth}
        \centering
        \includegraphics[width=\linewidth]{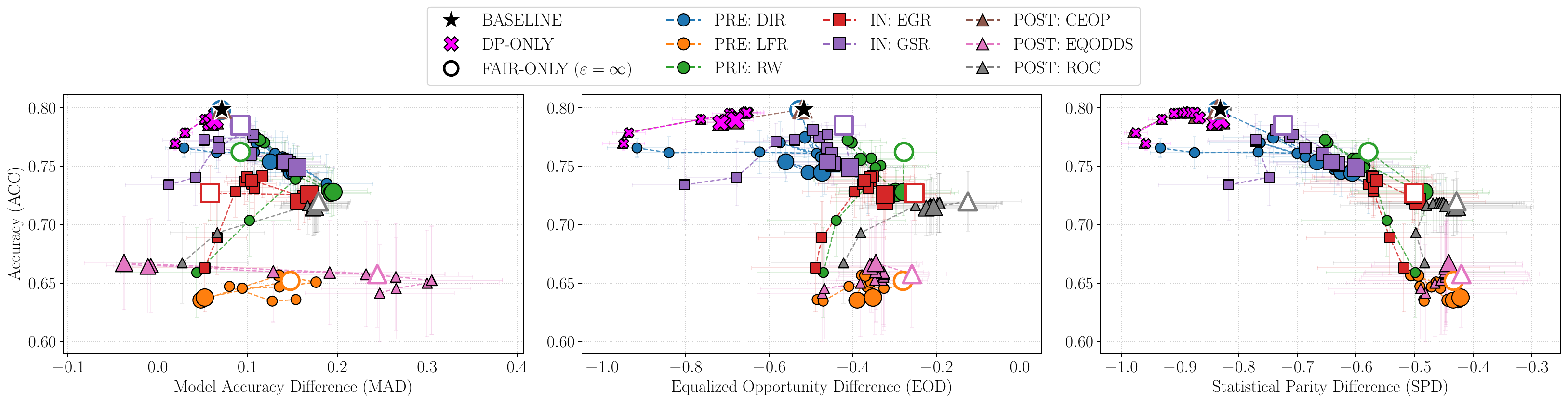}
        \caption{\textbf{BiasOnDemand Configuration 4}.}
        \label{fig-app:results-pareto-acc-bod-4-lr}
    \end{subfigure}\\
    \hfill
    \begin{subfigure}{0.99\textwidth}
        \centering
        \includegraphics[width=\linewidth]{figures/appendix/fig_results_LR_aim_ACC_all_BoD-5.pdf}
        \caption{\textbf{BiasOnDemand (Configuration 5)}.}
        \label{fig-app:results-pareto-acc-bod-5-lr}
    \end{subfigure}\\
    \begin{subfigure}{0.99\textwidth}
        \centering
        \includegraphics[width=\linewidth]{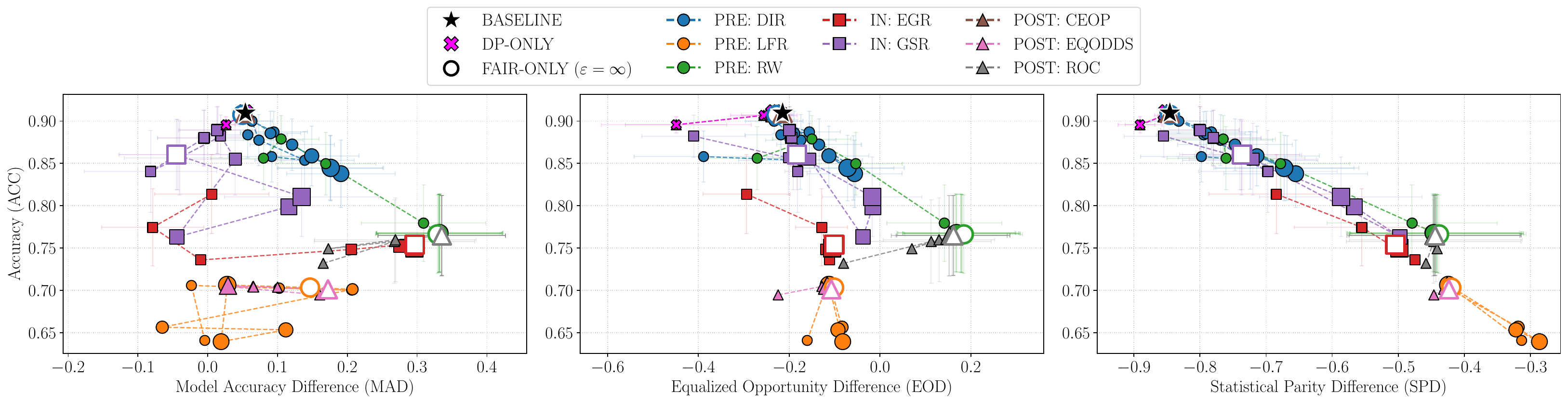}
        \caption{\textbf{BiasOnDemand Configuration 6}.}
        \label{fig-app:results-pareto-acc-bod-6-lr}
    \end{subfigure}\\
    \hfill
    \caption{Accuracy--fairness trade-off under the \textbf{AIM} synthesizer across configurations 4, 5, and 6 of BoD dataset with the classifier \textbf{Logistic Regression}. Each subfigure reports accuracy (ACC) versus three fairness metrics (MAD, EOD, SPD) for each fairness intervention applied at the pre-, in-, and post-processing stages. 
    Marker size encodes the privacy budget $\varepsilon$ for both \dponly{} and \dpfair{} settings, while hollow markers indicate the \faironly{} setting ($\varepsilon=\infty$).
    The \baseline{} setting (no DP and no fairness intervention) is indicated by the $\star$ symbol.}
    \label{fig-app:results-pareto-aim-acc-bod-456-lr}
\end{figure*}

\begin{figure*}[!ht]
    \centering
    \begin{subfigure}{0.99\textwidth}
        \centering
        \includegraphics[width=\linewidth]{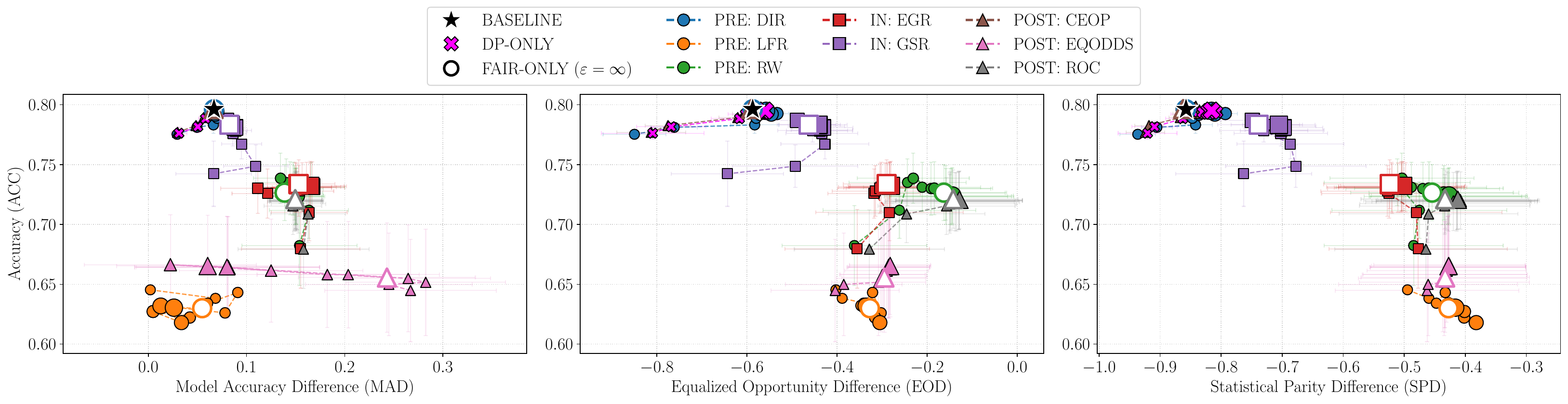}
        \caption{\textbf{BiasOnDemand Configuration 1}.}
        \label{fig-app:results-pareto-acc-bod-1-rf}
    \end{subfigure}\\
    \hfill
    \begin{subfigure}{0.99\textwidth}
        \centering
        \includegraphics[width=\linewidth]{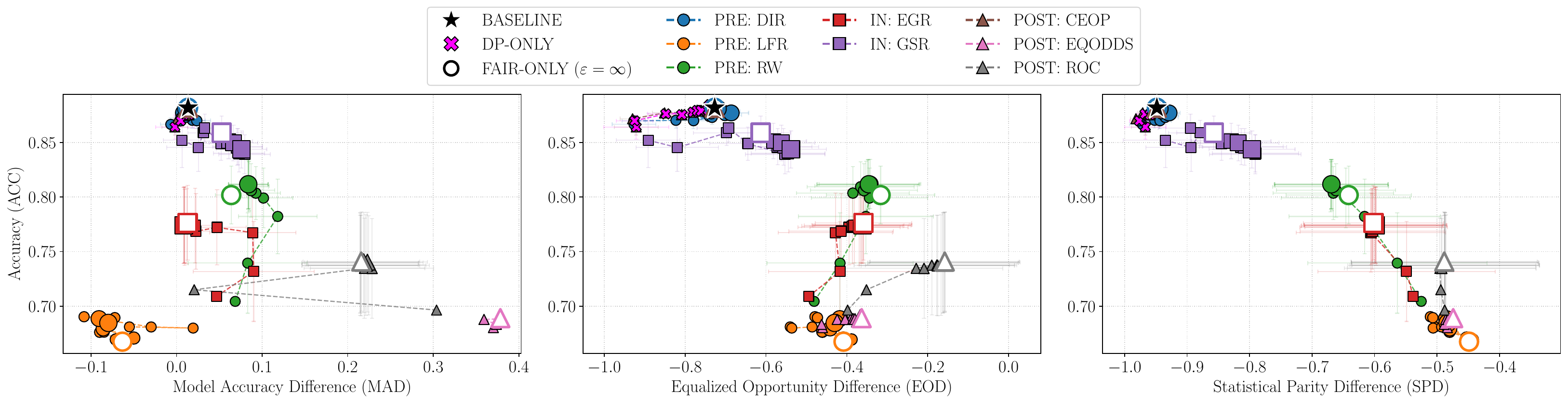}
        \caption{\textbf{BiasOnDemand Configuration 2}.}
        \label{fig-app:results-pareto-acc-bod-2-rf}
    \end{subfigure}\\
    \begin{subfigure}{0.99\textwidth}
        \centering
        \includegraphics[width=\linewidth]{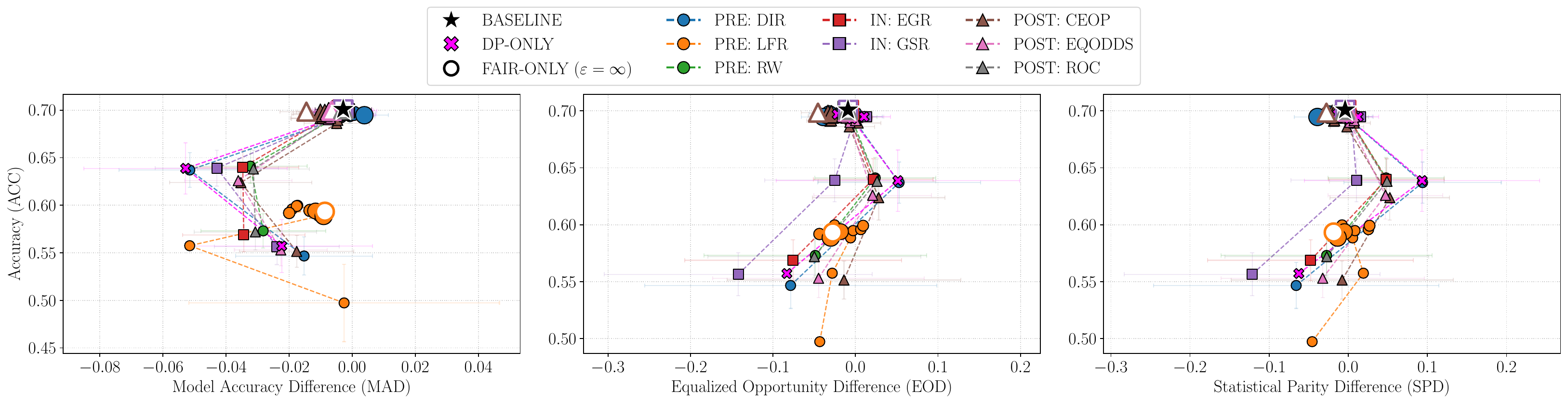}
        \caption{\textbf{BiasOnDemand Configuration 3}.}
        \label{fig-app:results-pareto-acc-bod-3-rf}
    \end{subfigure}\\
    \hfill
    \caption{Accuracy--fairness trade-off under the \textbf{AIM} synthesizer across configurations 1, 2, and 3 of the BoD dataset with the \textbf{Random Forest} classifier. Each subfigure reports accuracy (ACC) versus three fairness metrics (MAD, EOD, SPD) for each fairness intervention applied at the pre-, in-, and post-processing stages. 
    Marker size encodes the privacy budget $\varepsilon$ for both \dponly{} and \dpfair{} settings, while hollow markers indicate the \faironly{} setting ($\varepsilon=\infty$).
    The \baseline{} setting (no DP and no fairness intervention) is indicated by the $\star$ symbol.}
    \label{fig-app:results-pareto-aim-acc-bod-123-rf}
\end{figure*}

\begin{figure*}[!ht]
    \centering
    \begin{subfigure}{0.99\textwidth}
        \centering
        \includegraphics[width=\linewidth]{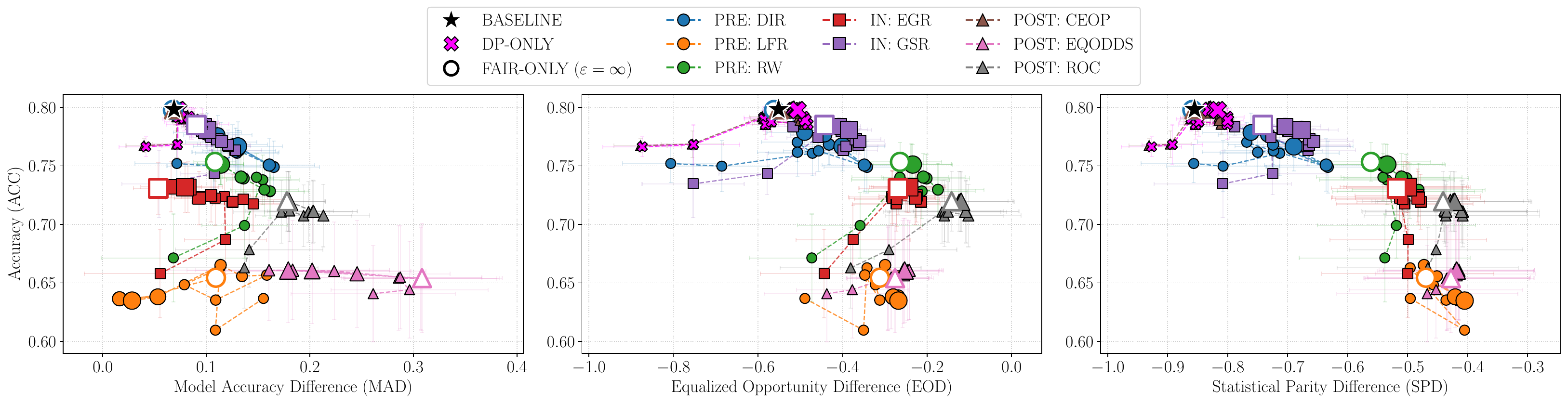}
        \caption{\textbf{BiasOnDemand Configuration 4}.}
        \label{fig-app:results-pareto-acc-bod-4-rf}
    \end{subfigure}\\
    \hfill
    \begin{subfigure}{0.99\textwidth}
        \centering
        \includegraphics[width=\linewidth]{figures/appendix/fig_results_RF_aim_ACC_all_BoD-5.pdf}
        \caption{\textbf{BiasOnDemand (Configuration 5)}.}
        \label{fig-app:results-pareto-acc-bod-5-rf}
    \end{subfigure}\\
    \begin{subfigure}{0.99\textwidth}
        \centering
        \includegraphics[width=\linewidth]{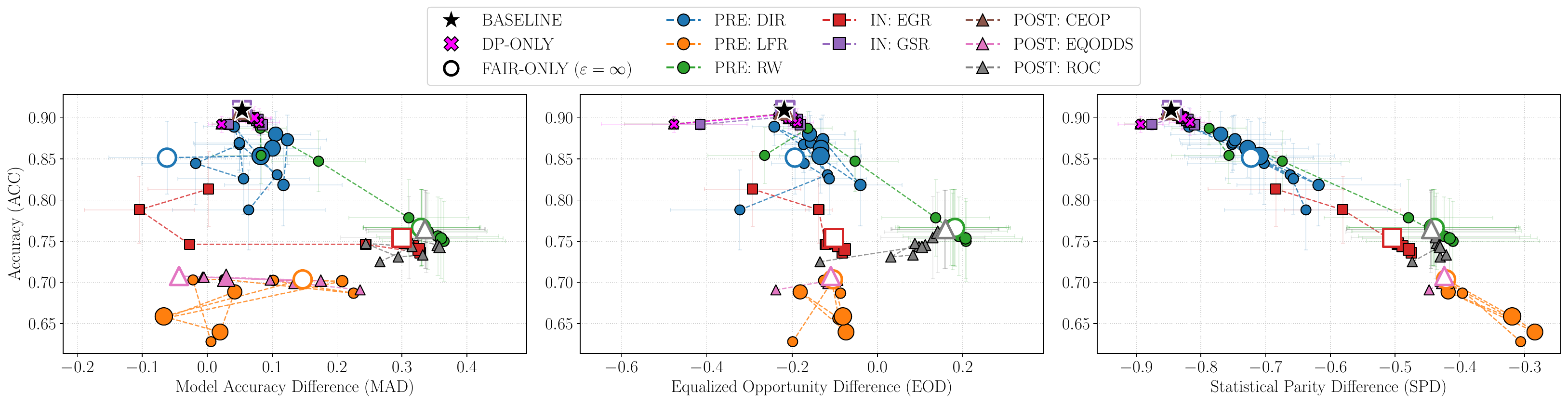}
        \caption{\textbf{BiasOnDemand Configuration 6}.}
        \label{fig-app:results-pareto-acc-bod-6-rf}
    \end{subfigure}\\
    \hfill
    \caption{Accuracy--fairness trade-off under the \textbf{AIM} synthesizer across configurations 4, 5, and 6 of the BoD dataset with the \textbf{Random Forest} classifier. Each subfigure reports accuracy (ACC) versus three fairness metrics (MAD, EOD, SPD) for each fairness intervention applied at the pre-, in-, and post-processing stages. 
    Marker size encodes the privacy budget $\varepsilon$ for both \dponly{} and \dpfair{} settings, while hollow markers indicate the \faironly{} setting ($\varepsilon=\infty$).
    The \baseline{} setting (no DP and no fairness intervention) is indicated by the $\star$ symbol.}
    \label{fig-app:results-pareto-aim-acc-bod-456-rf}
\end{figure*}

Figures~\ref{fig-app:results-pareto-mst-acc-bod-123-xgb}--\ref{fig-app:results-pareto-aim-acc-bod-456-rf} report the results obtained with XGBoost, Logistic Regression, and Random Forest classifiers across the six configurations of the Bias On Demand (BoD) dataset. Analyzing each fairness mechanism separately, several consistent patterns emerge. First, regardless of the induced bias type (imbalance, historical bias, or measurement bias), Reject Option Classification (ROC) consistently corrects group disparities, often bringing them close to zero, while keeping utility loss relatively bounded (on the order of one digit to low double-digit percentage points). Second, Equalized Odds post-processing (EqOdds) follows a similar correction pattern to ROC, but typically incurs a larger utility loss for comparable levels of bias mitigation. Third, Learning Fair Representations (LFR) exhibits the same behavior observed in other datasets: it effectively enforces parity across all bias types, but does so at the cost of the largest utility degradation among the evaluated mechanisms. Fourth, both Exponentiated Gradient Reduction (EGR) and Reweighing achieve comparable fairness--utility trade-offs overall; however, they are more sensitive to the specific type of induced bias and to the classifier used, as their relative performance varies across BoD configurations. Finally, the remaining mechanisms show some ability to reduce bias in specific settings, but their behavior is less stable, with either limited correction capability or high variability across bias types and classifiers.

% BOD - MST - XGB

% BOD - AIM - LR

% BOD - AIM - RF

\subsubsection{Correlation Between Classifier, Fairness Mechanisms and Dataset}
\label{app:correlation}
The results presented in this appendix may suggest a dependence of fairness mechanisms on the specific experimental configuration, \ie, the combination of classifier, fairness intervention, and dataset. While such correlations do exist, our analysis shows that they primarily affect the \emph{degree} of bias correction rather than its feasibility. Indeed, several mechanisms exhibit consistent qualitative behavior across classifiers and datasets, with only a few isolated cases deviating from the general patterns observed in the main results. This indicates that these mechanisms remain effective for correcting bias in models trained on \emph{differentially private} synthetic data, even when performance varies across configurations. In particular, the post-processing methods ROC and EqOdds stand out for their robustness, consistently achieving stable and well-balanced fairness--utility trade-offs across all experimental settings.

\end{document}